\documentclass{article}
\usepackage[preprint]{neurips_2026}
\usepackage[T1]{fontenc}
\usepackage[utf8]{inputenc}
\usepackage{amsmath,amssymb,amsthm,mathtools}
\DeclareMathOperator*{\argmin}{arg\,min}
\usepackage{booktabs,algorithm,algpseudocode}
\usepackage{hyperref,url,microtype}
\usepackage{wrapfig}
\hypersetup{colorlinks=true,linkcolor=blue,citecolor=blue,urlcolor=blue,
 pdftitle={EQUIP: Trajectory-Local Curvature Equilibration for Deep Policy Search}}
\newcommand{\method}{\textsc{Equip}}
\newcommand{\sacmethod}{\textsc{Equip-SAC}}
\newcommand{\R}{\mathbb{R}}
\newcommand{\E}{\mathbb{E}}
\newcommand{\tr}{\operatorname{tr}}
\newcommand{\diag}{\operatorname{diag}}
\newcommand{\clip}{\operatorname{clip}}
\newcommand{\KL}{D_{\mathrm{KL}}}
\newcommand{\TV}{D_{\mathrm{TV}}}
\newcommand{\norm}[1]{\left\lVert #1\right\rVert}
\newcommand{\ip}[2]{\left\langle #1,#2\right\rangle}
\newcommand{\loss}{\ell}
\newcommand{\epsm}{\varepsilon_{\mathrm m}}
\newcommand{\epss}{\varepsilon_{\mathrm s}}
\newtheorem{theorem}{Theorem}[section]
\newtheorem{proposition}[theorem]{Proposition}

\newtheorem{corollary}[theorem]{Corollary}
\theoremstyle{definition}
\newtheorem{assumption}[theorem]{Assumption}
\theoremstyle{remark}

\title{Curvature-Equilibrated Policy Optimization for Deep Reinforcement Learning}

\author{
Gang Chen\\
School of Engineering and Computer Science\\
Victoria University of Wellington\\
Wellington, New Zealand\\
\texttt{aaron.chen@ecs.vuw.ac.nz}
}

\begin{document}
\maketitle

\begin{abstract}
Gradient-based actor optimization is central to deep reinforcement
learning. However, conventional first-order optimizers do not explicitly
exploit the local curvature of neural actor losses.
We propose \method\ (\emph{Curvature-Equilibrated Gradients for Policy
Search}), a novel approach that seeks to balance local contraction
and expansion of the actor parameter space by equilibrating positive
and negative curvature. \method\ constructs a trajectory-informed
search subspace and uses the projected Hessian of actor loss to build
a bounded local metric for curvature equilibration.
First-order loss matching preserves the approximated
first-order decrease in actor loss. Meanwhile, gradient mixing and curvature-based
gradient selection adjust the gradient applied for actor optimization.
We establish the geometric properties of curvature equilibration,
derive bounds on gradient alignment and magnitude, and identify
conditions for improved actor loss reduction and convergence
toward stationarity under direct gradient updates.
We integrate \method\ into Soft Actor--Critic (SAC) without modifying
its actor loss, critic learning, or replay mechanism.
Experiments on 6 MuJoCo continuous control benchmarks show that
the resulting \sacmethod\ achieves higher mean sample efficiency
than SAC and two second-order policy optimization
baselines. Ablation studies further highlight the benefits of
curvature equilibration and trajectory-informed geometry for
policy training.
\end{abstract}

\section{Introduction}
\label{sec:intro}

Gradient-based policy optimization is fundamental to deep reinforcement
learning (DRL). Actor--critic algorithms, such as DDPG, TD3, and SAC,
optimize neural policies using actor-loss gradients guided by learned
action-value functions
\citep{lillicrap2016ddpg,fujimoto2018td3,haarnoja2018sac}.
Despite their success, actor optimization remains challenging because
neural actor losses are highly nonconvex and evolve as critics
and training data change. Common first-order actor optimizers do not
explicitly exploit the local curvature of the actor loss.
Yet, curvature information can help identify more promising actor
update directions~\citep{maniyar2024policynewton,khorasani2025vrscp},
offering valuable opportunities for greater loss reduction
and more effective policy optimization.

Geometry-aware and second-order methods seek to exploit such
information. For example, natural policy gradient uses the Fisher information
matrix to adjust policy updates \citep{kakade2001natural},
while Hessian-based methods use local curvature to construct
actor optimization directions
\citep{maniyar2024policynewton,kordabad2025compatible}.
However, curvature computations can be expensive
for neural actors with a large number of trainable parameters.
Moreover, negative or near-zero Hessian eigenvalues may cause
inverse-curvature updates to produce ascent directions or
excessively large actor parameter changes~\citep{dauphin2014identifying}.
These limitations demand for alternative approaches that exploit
curvature information efficiently and reliably.

A promising direction arises from the local geometry of
gradient-based actor optimization. For sufficiently small gradient
updates, positive and negative curvature contribute respectively to local
contraction and expansion of the actor parameter space~\citep{dauphin2014identifying,chen2023metricpg}.
While both effects can benefit optimization, excessive contraction
or expansion may undermine optimization performance and stability.

We take a new \emph{geometric perspective} by considering the balance
between these competing effects as a way to guide actor optimization.
Rather than suppressing curvature or approximating its inverse,
we investigate \emph{how equilibrating positive and negative curvature
can lead to more effective and reliable actor updates}.

Motivated by this geometric insight, we propose \method\
(\emph{Curvature-Equilibrated Gradients for Policy Search}),
a general mechanism for curvature-aware actor optimization.
\method\ constructs a \emph{low-dimensional policy search subspace}
from recent actor gradients.
It then estimates the projected Hessian using Hessian--vector
products and constructs a bounded local metric to equilibrate
positive and negative curvature,
enabling curvature-aware gradient adjustments without
computing or inverting the full Hessian. Periodically refreshing
and reusing the local geometry further reduces computational overhead.

Since curvature equilibration alone does not guarantee improved
actor updates, \method\ further introduces first-order loss matching,
gradient mixing, and curvature-based gradient selection techniques.
First-order loss matching preserves the approximated
first-order decrease in actor loss achievable by the raw gradient, while gradient
mixing controls the adjustment strength and bounds the mixed
gradient norm. Gradient selection then favors the mixed gradient
when the local curvature suggests a greater decrease
in actor loss than the raw gradient.
Together, these mechanisms exploit local curvature while
maintaining controlled actor updates.

We integrate \method\ into SAC~\citep{haarnoja2018sac}, resulting in \sacmethod.
\method\ modifies only the gradient supplied
to the actor optimizer, while preserving SAC's actor loss, critic
learning procedure, and replay mechanism. 
Experiments on 6 MuJoCo continuous-control benchmarks
demonstrate that \sacmethod\ achieves higher mean sample
efficiency and evaluation returns than SAC and two
second-order policy optimization baselines under the same
environment-interaction budget. Ablation studies further
support the effectiveness of the key algorithmic components,
while parameter sensitivity analyses provide additional
evidence for the integrated design of \method.

The main contributions of this work are summarized as follows.

\begin{itemize}
 \item We propose \method\ as a curvature-equilibration mechanism for
 gradient-based policy optimization. It constructs a bounded
 local metric from the projected actor-loss Hessian within
 a search subspace formed from recent actor gradients,
 enabling efficient use of positive and negative curvature
 without computing or inverting the full Hessian.

 \item We develop a gradient adjustment mechanism combining
 first-order loss matching, gradient mixing, and curvature-based
 selection. The resulting policy updates preserve the predicted
 first-order decrease in actor loss, control gradient magnitude,
 and can achieve greater actor loss reduction than raw-gradient
 updates under suitable conditions.

 \item We establish theoretical properties of \method,
 including the geometric interpretation and feasibility
 of curvature equilibration, gradient alignment and norm
 bounds, and conditions for improved actor optimization.
 We further derive stationarity guarantees for direct
 actor updates under fixed and changing actor losses,
 and relate actor-loss stationarity to the
 entropy-regularized return under bounded approximation errors.
\end{itemize}

\section{Related Work}
\label{sec:related_work}

We review related research in four areas: actor--critic
policy optimization, policy geometry and second-order
methods, efficient curvature estimation and gradient
subspaces, and geometry-aware policy optimization approaches.
We then summarize the key differences between \method\
and existing methods.

\paragraph{Actor--critic policy optimization.}
Policy-gradient methods optimize policies using gradients
of the expected return
\citep{sutton1999policygradient,silver2014deterministic}.
Actor--critic methods employ learned value functions to guide
actor optimization. DDPG and TD3 optimize deterministic
policies through critic-based action gradients
\citep{lillicrap2016ddpg,fujimoto2018td3}, while SAC
learns stochastic policies through an entropy-regularized
actor loss \citep{haarnoja2018sac}.

Recent advances in off-policy reinforcement learning
focus on improving value estimation, training efficiency,
and neural network architectures. REDQ combines critic
ensembles with frequent updates \citep{chen2021redq}.
CrossQ uses batch normalization to simplify critic
learning without target networks \citep{bhatt2024crossq},
while SimBa employs normalization and residual connections
to support larger networks \citep{lee2025simba}.

\paragraph{Policy geometry and second-order optimization.}
Natural policy gradient adjusts policy updates using
the inverse Fisher information matrix
\citep{kakade2001natural}. TRPO constrains policy
updates through KL divergence, while PPO uses a clipped
probability-ratio objective to limit policy changes
\citep{schulman2015trpo,schulman2017ppo}.
ACKTR approximates natural-gradient updates using
Kronecker-factored curvature information
\citep{wu2017acktr}. These methods primarily exploit
the geometry of policy distributions or constrain
policy changes.

Another line of research uses Hessian information
to construct second-order policy updates.
\citet{maniyar2024policynewton} estimate policy
gradients and Hessians from sampled trajectories
and employ cubic-regularized Newton updates to
escape saddle points. \citet{khorasani2025vrscp}
propose VR-SCP, which combines Hessian--vector products,
variance reduction, and cubic-regularized optimization
to improve the sample efficiency of reaching
second-order stationary points.
\citet{sun2025vrcrpn} further develop a variance-reduced
cubic-regularized Newton method based on Hessian
estimation of the expected return. Recent studies
also investigate second-order optimization for
actor--critic learning
\citep{kordabad2025compatible,manivannan2026secondorder}.

Among these methods, VR-SCP provides a recent
second-order policy optimization approach with
a publicly available implementation. We therefore
include it in our experimental comparisons.

\paragraph{Efficient curvature estimation and gradient subspaces.}
Several optimization methods use curvature estimates
or structured gradient statistics to improve training
efficiency. Sophia periodically estimates diagonal
curvature and clips its updates \citep{liu2024sophia},
while SOAP applies Adam \citep{kingma2015adam}
in a rotated coordinate system derived from
Shampoo preconditioning \citep{vyas2025soap}.
In deep reinforcement learning,
\citet{creuscastanyer2025stable} investigate
gradient degradation in large networks and
improve learning using residual connections
and structured preconditioning.

Low-dimensional curvature approximations offer
another way to reduce computational costs.
SketchySGD uses randomized curvature estimates
to construct preconditioners
\citep{frangella2024sketchysgd}, while CAO
periodically updates a low-rank Hessian
approximation for damped inverse-curvature scaling
\citep{du2025cao}.
For policy optimization,
\citet{schneider2024subspaces} find that PPO and
SAC gradients often concentrate in low-dimensional
subspaces spanned by leading Hessian eigenvectors,
which can remain similar across multiple policy updates.

\paragraph{Geometry-aware policy optimization approaches.}
Sharpness-aware minimization (SAM) seeks parameters
with low loss under nearby parameter perturbations
\citep{foret2021sam}. In reinforcement learning,
Q-SAM extends sharpness-aware optimization to offline
policy learning to improve generalization of trained
policies~\citep{wang2026qsam}.

Another approach uses a local metric to control
policy-search geometry. Specifically, \citet{chen2023metricpg}
propose a method that trains a dedicated metric tensor network
to reduce the absolute divergence of the policy-gradient
field. They demonstrate its effectiveness with SAC and TD3.

\paragraph{Key distinctions and contributions of \method.}
Existing studies provide several ways to exploit
policy geometry and curvature information, including
natural-gradient updates, Newton-type optimization,
curvature-based preconditioning, and sharpness-aware
learning. Compared to these works, \method\ introduces
a different approach that explicitly equilibrates
positive and negative curvature within a
trajectory-informed search subspace.

Unlike Newton-type methods that use curvature to
construct second-order update directions, \method\
directly constructs a bounded local metric from the
projected actor-loss Hessian to adjust actor gradients.
Unlike metric-based approaches that train an additional
network, metric is computed directly in \method\ from local
curvature information. \method\ further
integrates first-order actor loss matching, gradient mixing,
and curvature-based gradient selection to preserve
the predicted first-order decrease in actor loss
and control gradient adjustments. These components
can be incorporated into many existing actor--critic
algorithms without modifying their policy optimization
objectives or critic learning procedures.

Furthermore, our theoretical analysis establishes the feasibility
of curvature equilibration under bounded metric weights
and provides bounds on gradient alignment and magnitude.
It also identifies conditions under which mixed gradients
achieve greater actor loss reduction than raw gradients.
We further derive convergence guarantees toward stationarity
under direct actor updates for fixed actor losses,
along with conditional stationarity bounds for changing
actor losses.

\section{Reinforcement Learning and Policy Search Preliminaries}
\label{sec:prelim}

\subsection{Problem formulation}
We consider an agent interacting with an environment described by a
discounted \emph{Markov decision process} (MDP),
$\mathcal M=(\mathcal S,\mathcal A,P,R,\rho_0,\gamma)$.
Here, $\mathcal S$ and $\mathcal A$ are the state and action spaces.
The distribution $P(\cdot\mid s,a)$ describes the next state after taking
action $a$ in state $s$, and $R(s,a)$ is the expected bounded reward.
The initial state follows $\rho_0$, and $\gamma\in(0,1)$ is the discount factor.

The agent follows a differentiable neural policy $\pi_\theta$ as its actor, which is
parameterized by $\theta\in\R^p$, where $p$ is the number of trainable parameters in
the neural actor.
Starting from $s_0\sim\rho_0$, it samples
$a_k\sim\pi_\theta(\cdot\mid s_k)$ and then observes
$s_{k+1}\sim P(\cdot\mid s_k,a_k)$. Policy search aims to find actor parameters
that maximize the \emph{expected discounted return},
\begin{equation}
 J(\theta)=\E_{\pi_\theta,P,\rho_0}
 \left[\sum_{k=0}^{\infty}\gamma^k R(s_k,a_k)\right].
 \label{eq:rl-objective}
\end{equation}
For a policy $\pi$, the state value $V^\pi(s)$ is the expected return
starting from $s$. The action value $Q^\pi(s,a)$ is the expected return
after taking $a$ in $s$ and following $\pi$ thereafter:
\begin{equation*}
 \begin{aligned}
 V^\pi(s)=\E_{\pi,P}
 \left[\sum_{k=0}^{\infty}\gamma^k R(s_k,a_k)\mid s_0=s\right], 
 Q^\pi(s,a)=R(s,a)+\gamma\E_{s'\sim P(\cdot\mid s,a)}V^\pi(s').
 \end{aligned}
\end{equation*}
We summarize the states visited by $\pi_\theta$ using its normalized
discounted state distribution,
\begin{equation}
 d_\theta(s)=(1-\gamma)\sum_{k=0}^{\infty}\gamma^k
 p_{\theta,k}(s),
 \label{eq:occupancy}
\end{equation}
where $p_{\theta,k}(s)$ is the probability mass or density of the state at
step $k$. The factor $1-\gamma$ makes $d_\theta$ a probability distribution.
This distribution determines how states are weighted in the
policy-gradient formula in Equation~\eqref{eq:pg} below.

We also consider entropy-regularized policy search, which encourages
stochastic policies by adding an entropy term to the return:
\begin{equation}
 J_\alpha(\theta)=\E_{\pi_\theta,P,\rho_0}
 \left[\sum_{k=0}^{\infty}\gamma^k
 \bigl(R(s_k,a_k)-\alpha\log\pi_\theta(a_k\mid s_k)\bigr)\right].
 \label{eq:soft-objective}
\end{equation}
The coefficient $\alpha>0$ controls the balance between environmental reward
and policy entropy. For continuous actions, $\pi(a\mid s)$ denotes the policy's
probability density. The corresponding soft value functions are
\begin{equation}
 \begin{aligned}
 Q_\alpha^\pi(s,a)&=R(s,a)
 +\gamma\E_{s'\sim P(\cdot\mid s,a)}V_\alpha^\pi(s'),\\
 V_\alpha^\pi(s)&=\E_{a\sim\pi(\cdot\mid s)}
 [Q_\alpha^\pi(s,a)-\alpha\log\pi(a\mid s)].
 \end{aligned}
 \label{eq:soft-values}
\end{equation}
Thus, $Q_\alpha^\pi$ includes the immediate reward and future soft value,
while $V_\alpha^\pi$ also includes the action-selection entropy of the current state.
These definitions are used in the SAC critic target in
Equation~\eqref{eq:sac-target} and the actor loss in
Equation~\eqref{eq:sac-actor} in Section~\ref{sec:sac}.

In this work, we aim to improve gradient-based policy search through \method\ as a general
mechanism, with SAC as one application. We use $k$ for environment steps
and $t$ for actor/policy updates. These counters can advance at different rates
because actor updates can reuse data from a replay buffer. 

\subsection{From policy gradients to actor--critic optimization}

The policy-gradient theorem expresses the gradient of the
expected return with respect to the actor parameters
\citep{sutton1999policygradient}:
\begin{equation}
 \nabla_\theta J(\theta)=\frac{1}{1-\gamma}
 \E_{s\sim d_\theta,\,a\sim\pi_\theta}
 [\nabla_\theta\log\pi_\theta(a\mid s)Q^{\pi_\theta}(s,a)].
 \label{eq:pg}
\end{equation}
This provides the foundation for gradient-based policy
optimization. In actor--critic methods, a learned critic
estimates action values or advantages to guide actor updates.
Different algorithms construct their actor losses differently.
For example, DDPG and TD3 optimize deterministic policies
using critic-based action gradients
\citep{lillicrap2016ddpg,fujimoto2018td3}, while PPO uses
a clipped policy-gradient objective \citep{schulman2017ppo}.
SAC combines critic-based policy optimization with entropy
regularization \citep{haarnoja2018sac}.

Despite these differences, gradient-based actor--critic methods
share a common optimization process: they \emph{update
actor parameters using gradients of an actor loss}.
\method\ builds on this process by incorporating local curvature
information to adjust the actor gradient without changing the
underlying actor loss or critic learning procedure.
This makes \method\ potentially applicable to different
actor--critic algorithms, provided that their actor losses support
the required gradient and curvature computations.
In this study, we focus on SAC as a representative application.

\subsection{Why apply \method\ to SAC?}

SAC combines entropy-regularized policy optimization with
off-policy actor--critic learning
\citep{haarnoja2018sac,haarnoja2018sacapplications}.
It provides a suitable framework for \method\ because its
actor loss is directly related to policy improvement and
supports the gradient and curvature computations required
by our method.

To establish this connection, consider a fixed state $s$,
an action-value function $q$, and an entropy temperature
$\alpha>0$. Define the Gibbs policy
\[
 p_q(a\mid s)=\frac{\exp(q(s,a)/\alpha)}{Z_q(s)},
 \qquad
 Z_q(s)=\int_{\mathcal A}\exp(q(s,a)/\alpha)\,d\nu(a),
\]
where $\nu$ is a fixed reference measure and $Z_q(s)$
is assumed finite. The entropy-regularized actor loss
at state $s$ can be expressed as
\begin{equation}
 \E_{a\sim\pi}
 [\alpha\log\pi(a\mid s)-q(s,a)]
 =\alpha\KL(\pi(\cdot\mid s)\Vert p_q(\cdot\mid s))
 -\alpha\log Z_q(s).
 \label{eq:gibbs-kl}
\end{equation}
Since $Z_q(s)$ does not depend on $\pi$, minimizing the
actor loss is equivalent to minimizing the KL divergence
between the current policy $\pi$ and the Gibbs policy $p_q$.
This provides a direct connection between actor
loss minimization and entropy-regularized policy
improvement. Specifically, we extend the soft policy-improvement
argument of~\citet[Lemma~2]{haarnoja2018sac} to account for
critic approximation and imperfect actor optimization,
as formalized below.

\begin{proposition}[Exact and approximate soft policy improvement]
\label{prop:soft-improvement}
Let $\pi$ and $\pi'$ denote the current and updated policies,
respectively, and let $q=Q_\alpha^\pi$ be the exact soft
action-value function of $\pi$. Assume the soft value functions
and entropy expectations are uniformly bounded.
Suppose an estimated critic $\widehat q$ satisfies
$\norm{\widehat q-q}_\infty\leq\epsilon_Q$ and, for every state $s$,
\[
 \E_{a\sim\pi'}[
 \widehat q(s,a)-\alpha\log\pi'(a\mid s)]
 \geq
 \E_{a\sim\pi}[
 \widehat q(s,a)-\alpha\log\pi(a\mid s)]
 -\epsilon_{\rm opt},
\]
where $\epsilon_{\rm opt}\geq0$.
Then, for every state $s$,
\begin{equation}
 V_\alpha^{\pi'}(s)\geq V_\alpha^\pi(s)
 -\frac{2\epsilon_Q+\epsilon_{\rm opt}}{1-\gamma}.
 \label{eq:approx-soft-improvement}
\end{equation}
In particular, with an exact critic ($\epsilon_Q=0$)
and no loss of improvement ($\epsilon_{\rm opt}=0$),
the updated policy satisfies
$V_\alpha^{\pi'}(s)\geq V_\alpha^\pi(s)$.
\end{proposition}

The proof is provided in Appendix~\ref{app:soft-proof}.
Proposition~\ref{prop:soft-improvement} shows that reducing
the actor loss can improve the policy's soft value, while
critic approximation and imperfect actor optimization
limit this guarantee. This motivates applying \method\
to enhance the optimization of the actor loss in SAC. 
Proposition~\ref{prop:soft-improvement} establishes the connection
between actor loss minimization and policy improvement under a
state-wise condition. Since \method\ operates on sampled actor
losses, we separately analyze its gradient adjustments and
optimization properties in Section~\ref{sec:theory}.

SAC is also well suited to the implementation and
evaluation of \method. Its reparameterized actor loss
allows gradients and Hessian--vector products to be
computed using the same replay minibatch, actor noise,
and frozen critics. Moreover, \method\ modifies only
the actor gradient, preserving SAC's critic learning
procedure and replay mechanism. This enables a controlled
comparison between SAC and \sacmethod\ (see Section~\ref{sec:exp-main}) under the same
environment interaction budgets.

\section{\method: Curvature-Equilibrated Gradients for Policy Search}
\label{sec:method}

\subsection{Algorithm overview}

We introduce \method, (\emph{Curvature-Equilibrated
Gradients for Policy Search}), a general mechanism for curvature-aware
actor optimization. The central idea is to construct a \emph{bounded local
metric} that equilibrates positive and negative curvature within a search
subspace, followed by gradient rescaling and mixing to approximately
preserve the predicted first-order decrease in actor loss.
This curvature-equilibration approach is applicable to a broad range of
gradient-based policy optimization algorithms, with its integration
into SAC detailed in Section~\ref{sec:sac}.

For any base actor-critic algorithm such as SAC, let the current actor update
be $t$. The base algorithm provides a differentiable actor loss $\loss_t$. Its \emph{raw gradient}
$g_t$ and \emph{Hessian} $H_t$ are
\begin{equation}
 \loss_t(\theta)=\loss(\theta;\mathcal B_t,\xi_t,\zeta_t),\qquad
 g_t=\nabla_\theta\loss_t(\theta_t),\quad
 H_t=\nabla_\theta^2\loss_t(\theta_t),
 \label{eq:interface}
\end{equation}
where $\mathcal B_t$ denotes the data batch and $\xi_t$ represents the sampled actor noise.
The term $\zeta_t$ collects additional quantities required by the base algorithm, such as
critic parameters, advantage estimates, or reference-policy outputs. With $\mathcal B_t$, $\xi_t$,
and $\zeta_t$ fixed, $H_t$ measures the
curvature of the actor loss $\loss_t(\theta)$.

To make curvature calculations tractable for neural actors, \method\ 
constructs a \emph{low-dimensional policy search subspace} from the current raw gradient
$g_t$ as well as recent gradients utilized for actor optimization
(Subsection~\ref{sec:search-subspace}). Within this subspace,
it constructs a bounded local metric to reduce the imbalance between
positive and negative curvature contributions
(Subsection~\ref{sec:curvature-balancing};
Equation~\eqref{eq:practical-metric}).
The metric-adjusted gradient is then rescaled to match the approximated first-order
decrease in actor loss achieved by a direct raw-gradient update
(Subsection~\ref{sec:rescaling-mixing};
Equation~\eqref{eq:matching-scale}) and mixed with $g_t$.
A gradient-selection mechanism (Subsection~\ref{sec:gradient-selection})
chooses between the mixed and raw gradients, yielding the final gradient
$u_t$ as input to the actor optimizer $\mathcal O$. We use Adam~\citep{kingma2015adam} as $\mathcal O$ in our
experiments, while the theoretical analysis in Section~\ref{sec:theory}
considers the direct update $\theta_{t+1}=\theta_t-\eta_tu_t$ for simplicity.
Algorithm~\ref{alg:mechanism} summarizes the complete procedure.
Subsection~\ref{sec:computational-cost} further analyzes \method's computational and
memory complexity.

\subsection{Forming a search subspace from recent gradient directions}
\label{sec:search-subspace}

Forming the full $p\times p$ Hessian of the actor loss is computationally
prohibitive. Instead, \method\ estimates curvature within a low-dimensional
search subspace constructed from recent actor gradients.

At actor update $t$, we retain the gradient vectors used in the previous
$h$ completed actor updates and combine them with the current raw gradient
$g_t$. Including $g_t$ ensures that the search subspace captures the
current gradient direction, which may not be adequately represented by
historical gradients. We normalize these vectors and apply Gram--Schmidt
orthogonalization to remove linearly dependent directions, followed by
a reduced QR decomposition to improve numerical orthonormality.
The resulting $r_t$ orthonormal vectors $q_j$ ($1\leq j\leq r_t$) form
the basis $Q_t=[q_1,\ldots,q_{r_t}]\in\R^{p\times r_t}$ of the search subspace.
If $r_t<r_{\min}$, the subspace is not sufficiently ranked, and the actor
is updated using the raw gradient $g_t$.

When $r_t\geq r_{\min}$, we compute a \emph{Hessian--vector product}
(HVP) $H_tq_j$ for each basis vector $q_j$, requiring $r_t$ HVPs in total.
All HVPs are evaluated using the same actor loss, with the same batch
$\mathcal B_t$, actor noise $\xi_t$, and auxiliary quantities $\zeta_t$
held fixed. The HVPs and the resulting projected Hessian $A_t$ are computed as
\begin{equation}
 \begin{aligned}
 H_tq_j&=\left.\nabla_\theta
 \left[(\nabla_\theta\loss_t(\theta))^\top q_j\right]
 \right|_{\theta=\theta_t},\\
 A_t&=\frac12\bigl(Q_t^\top H_tQ_t+(Q_t^\top H_tQ_t)^\top\bigr)
      =U_t\diag(\mu_t)U_t^\top.
 \end{aligned}
 \label{eq:projected-hessian}
\end{equation}
By treating the basis vectors as fixed during differentiation, these
HVPs can be computed without explicitly forming the full Hessian
\citep{pearlmutter1994hvp}. We symmetrize the projected Hessian to
eliminate numerical asymmetry and compute its eigendecomposition in
double precision. Each eigenvalue $\mu_{t,i}$ represents the curvature of
the actor loss along a direction within the search subspace,
identified by the corresponding column of $Q_tU_t$. Positive eigenvalues indicate
\emph{upward curvature}, while negative eigenvalues indicate \emph{downward curvature}.

\subsection{Local curvature equilibration with a non-Euclidean metric}
\label{sec:curvature-balancing}

A \emph{local metric} determines how actor parameter changes
are measured within the search subspace. Any such change can
be expressed as $Q_tv$, where $v\in\R^{r_t}$ contains its
coordinates in the orthonormal basis $Q_t$.
A positive-definite metric $G\in\R^{r_t\times r_t}$
defines the squared length of this change as
\[
 \norm{v}_G^2=v^\top Gv.
\]
The Euclidean metric $G=I_{r_t}$ treats all directions within
the search subspace equally, whereas a non-Euclidean metric
assigns different weights to different directions.
Let $B=G^{-1}$ denote the \emph{inverse metric}. For any raw actor
gradient $g_t\in\R^p$, the corresponding metric-adjusted gradient
in the original actor parameter space is
$
 (I_p-Q_tQ_t^\top)g_t+Q_tBQ_t^\top g_t.
$
This adjustment reweights the gradient within the search subspace
while preserving its component outside the subspace.

Let $A_t=U_t\diag(\mu_t)U_t^\top$ be the projected Hessian
in Equation~\eqref{eq:projected-hessian}. We construct the
inverse metric $B_t$ using the same eigenvectors as $A_t$,
with a positive weight $b_{t,i}$ assigned to each eigenvector:
\begin{equation}
 B_t=U_t\diag(b_t)U_t^\top,\qquad
 b_{t,i}\in[m,M],\quad 0<m\leq1\leq M.
 \label{eq:metric}
\end{equation}
Under the local metric $G_t=B_t^{-1}$, the $i$-th
eigenvector of $A_t$ has a squared metric length of
$1/b_{t,i}$. Since $B_t$ scales the gradient component
along this eigenvector by $b_{t,i}$, larger weights
increase its contribution to the metric-adjusted gradient,
while smaller weights reduce it. Bounding the weights
within $[m,M]$ ensures that $B_t$ remains \emph{positive definite}
and limits gradient rescaling.

The curvature along the $i$-th eigenvector, measured
per unit squared length under $G_t$, is therefore
$b_{t,i}\mu_{t,i}$. Since $A_t$ and $B_t$
share the same eigenvectors, we have
\[
 B_tA_t=U_t\diag(b_{t,1}\mu_{t,1},\ldots,
 b_{t,r_t}\mu_{t,r_t})U_t^\top.
\]
The trace of $B_tA_t$ sums these signed curvature
contributions under $G_t$. As established in
Proposition~\ref{prop:geometry}, this \emph{metric-scaled trace} determines the
first-order local volume change under
a metric-adjusted gradient update. A positive trace indicates
local contraction, while a negative trace indicates expansion.
Although both effects can benefit optimization, excessive
contraction may restrict exploration of the actor loss landscape,
whereas excessive expansion may undermine optimization
stability. To balance these competing effects, \method\
aims to \emph{equilibrate} positive and negative curvature
contributions by targeting
\begin{equation}
 \tr(B_tA_t)=\sum_i b_{t,i}\mu_{t,i}=0.
 \label{eq:metric-target}
\end{equation}
To approach this \emph{curvature-equilibration target}, we compute
\begin{equation}
 \kappa_t=\frac{\sum_i\mu_{t,i}}{\sum_i\mu_{t,i}^2+\epsm},\qquad
 b_{t,i}=\clip(1-\kappa_t\mu_{t,i},m,M),
 \label{eq:practical-metric}
\end{equation}
where $\epsm>0$ prevents division by zero, and $\clip$ bounds the weights
within $[m,M]$\footnote{We use $m=1/3$ and $M=3$ in our experiments and
evaluate their impact on reinforcement learning performance.}.
When the metric-scaled trace is positive, this rule reduces the weights
on positive-curvature directions and increases those on negative-curvature
directions. The reverse occurs when the trace is negative.

Without regularization or weight bounds, Equation~\eqref{eq:practical-metric}
achieves exact curvature equilibration with the smallest squared deviation from
the unit weights $b_{t,i}=1$. Regularization and clipping may prevent
exact equilibration, as further analyzed in Proposition~\ref{prop:feasibility}.
We therefore use $B_t$ only when both curvature signs are present and
when the metric-scaled trace is sufficiently reduced:
\begin{equation}
 \max_i\mu_{t,i}>\tau_\mu,\quad
 \min_i\mu_{t,i}<-\tau_\mu,\qquad
 \frac{|\sum_i b_{t,i}\mu_{t,i}|}
 {|\sum_i\mu_{t,i}|+\epsm}<\rho_{\rm tr},
 \label{eq:metric-eligibility}
\end{equation}
where $\tau_\mu=\max(10^{-8},10^{-5}\max_i|\mu_{t,i}|)$ and
$\rho_{\rm tr}=0.9$ in our experiments. The first two conditions in Equation~\eqref{eq:metric-eligibility}
require positive and negative curvature beyond numerical tolerance, while the
last requires a reduction in trace imbalance relative to the Euclidean
metric. If any condition fails, we update the actor using the raw
gradient $g_t$.

Importantly, equilibrating the overall curvature does not necessarily
reduce curvature along the final actor update direction. We therefore
compare the raw and adjusted gradients within the same search subspace
and accept the adjustment only when it lowers the overall curvature
in the local quadratic loss model
(see Equation~\eqref{eq:curvature-margin}).

\subsection{First-order actor loss matching and gradient mixing}
\label{sec:rescaling-mixing}

The local metric can change the magnitude and direction of the metric-adjusted
gradient. To preserve the first-order decrease in actor loss, we must rescale the
metric-adjusted gradient.

Let $Q$ be the current search basis, $B$ the inverse metric, and
$z_t=Q^\top g_t$ the raw gradient in subspace coordinates. For
$z_t\neq0$, we define the scaling factor $\chi_t$ and the
\emph{adjusted gradient} $d_t$ as
\begin{equation}
 \chi_t=\frac{\norm{z_t}^2}{z_t^\top Bz_t},\qquad
 d_t=(I-QQ^\top)g_t+\chi_tQBz_t.
 \label{eq:matching-scale}
\end{equation}
The adjusted gradient $d_t$ preserves the gradient component of $g_t$ outside the search
subspace while reweighting and rescaling the component within it.
If $z_t=0$, we set $d_t=g_t$.

The scaling factor $\chi_t$ ensures that
$
 g_t^\top d_t=\norm{g_t}^2.
$
Consequently, for the same step size $\eta_t$, direct actor updates
along $-d_t$ and $-g_t$ both yield an expected first-order decrease
in actor loss of $\eta_t\norm{g_t}^2$.

We then mix the adjusted and raw gradients to obtain
\begin{equation}
 \widetilde u_t=(1-\lambda)g_t+\lambda d_t,
 \qquad \lambda\in[0,1].
 \label{eq:direction}
\end{equation}
The mixing weight $\lambda$ controls the strength of the curvature
adjustment, from the raw gradient ($\lambda=0$) to the fully adjusted
gradient ($\lambda=1$). Since
$g_t^\top\widetilde u_t=\norm{g_t}^2$, the mixed gradient also
preserves the first-order decrease in actor loss.
Theorem~\ref{thm:matching} establishes these properties and quantifies
the matching error introduced by numerical regularization.
Implementation details are provided in
Appendix~\ref{app:matching-implementation}.

\subsection{Gradient selection}
\label{sec:gradient-selection}

\method\ selects the mixed gradient to update the actor based on its projected curvature.
Let $z_t=Q^\top g_t$ and $\widetilde z_t=Q^\top\widetilde u_t$ denote
the subspace coordinates of the raw and mixed gradients, respectively,
and let $A$ be the most recently computed projected Hessian.
We compare their curvature terms:
\begin{equation}
 K_t^{\rm raw}=z_t^\top Az_t,\qquad
 K_t^{\rm mix}=\widetilde z_t^\top A\widetilde z_t,\qquad
 K_t^{\rm mix}<K_t^{\rm raw}.
 \label{eq:curvature-gate}
\end{equation}
Since the two gradients have matched first-order decrease in actor loss, this
condition favors the mixed gradient when it has a smaller quadratic
term in the projected actor loss. We accept the mixed gradient only
if this condition is satisfied. Otherwise, we use the raw gradient.
The final input to the actor optimizer $\mathcal{O}$ is therefore
\begin{equation}
 u_t=
 \begin{cases}
  \widetilde u_t, & \text{if the adjustment is accepted},\\
  g_t, & \text{otherwise}.
 \end{cases}
 \label{eq:accepted-input}
\end{equation}
Proposition~\ref{prop:cached-gap} establishes conditions under which
the projected curvature comparison in Equation~\eqref{eq:curvature-gate}
also guarantees a lower Hessian quadratic form for the mixed gradient
under the current actor loss in the full actor parameter space.
Detailed numerical acceptance criteria are provided in
Appendix~\ref{app:acceptance-implementation}.

\begin{algorithm}[t]
\caption{\method: curvature-equilibrated gradients for policy search}
\label{alg:mechanism}
\begin{algorithmic}[1]
\Require Actor loss $\loss_t$, parameters $\theta_t$, gradient history
$\mathcal H_t$, stored geometry $(Q,A,B)$, base optimizer $\mathcal O$,
refresh interval $K$, mixing weight $\lambda$
\State Compute the raw gradient $g_t=\nabla_\theta\loss_t(\theta_t)$
\State $u_t\gets g_t$
\If{a refresh is due (initially and every $K$ actor updates)}
 \State Build an orthonormal search basis $Q$ from recent gradients in $\mathcal H_t$ and $g_t$
 \If{the search subspace has sufficient rank}
  \State Compute the projected Hessian $A$ through Equation~\eqref{eq:projected-hessian}
  \State Construct the inverse metric $B$ with bounded weights through Equation~\eqref{eq:practical-metric}
 \EndIf
 \State Store the refreshed geometry; mark it unusable if any check in Equation~\eqref{eq:metric-eligibility} fails
\EndIf
\If{the stored geometry is usable}
 \State Re-weight and rescale the current gradient to obtain $d_t$ through Equation~\eqref{eq:matching-scale}
 \State Mix with the raw gradient: $\widetilde u_t\gets(1-\lambda)g_t+\lambda d_t$
 \State Compute the curvature terms $K_t^{\rm raw}$ and $K_t^{\rm mix}$ through Equation~\eqref{eq:curvature-gate}
 \If{$K_t^{\rm mix}<K_t^{\rm raw}$}
  \State $u_t\gets\widetilde u_t$
 \EndIf
\EndIf
\State Update the actor using $\mathcal O$ with gradient $u_t$
\State Store $u_t$ in the gradient history $\mathcal H_{t+1}$ and retain the most recent $h$ gradients
\State \Return $\theta_{t+1}$
\end{algorithmic}
\end{algorithm}

\subsection{Computational and memory complexity}
\label{sec:computational-cost}

\method\ refreshes the search subspace and local metric every $K$
actor updates, reusing them between refreshes. The mixed gradient is
still computed from the current raw gradient at each actor update.
Reusing the local geometry avoids repeated HVPs, although the cached
curvature may become outdated.
Proposition~\ref{prop:cached-gap} analyzes the effect of such changes,
and detailed refresh procedures are explained in
Appendix~\ref{app:geometry-reuse}.

Let $C_{\rm base}$ denote the cost of a standard actor update,
including gradient computation and parameter optimization, and let
$C_{\rm HVP}$ denote the cost of one HVP. For a search subspace of
dimension $r\leq r_{\max}\ll p$ ($p$ refers to the number of trainable
policy parameters), each refresh requires $r$ HVPs,
$\mathrm{O}(pr_{\max}^2)$ operations for basis construction and Hessian
projection (assuming $h=\mathrm{O}(r_{\max})$), and $\mathrm{O}(r^3)$ operations
for eigendecomposition. Computing and checking the mixed gradient
requires an additional $\mathrm{O}(pr)$ operations per actor update.
Since $r\leq r_{\max}\ll p$, the average computational cost per
actor update is
\begin{equation}
 \mathrm{O}\!\left(
 C_{\rm base}+pr+
 \frac{rC_{\rm HVP}+pr_{\max}^2}{K}
 \right).
 \label{eq:average-update-cost}
\end{equation}
The three terms represent the standard actor-update cost, the
per-update gradient adjustment cost, and the amortized geometry-refresh
cost, respectively. No additional HVPs are required between
refreshes. Furthermore, \method\ does not introduce additional environment
interactions. The additional memory required to store the gradient history,
search basis, and local geometry is $\mathrm{O}(ph+pr+r^2)$.

When $r\ll p$ and $h=\mathrm{O}(r)$, the additional memory scales
linearly with $p$, avoiding the quadratic storage cost of the full
Hessian. Moreover, the computational overhead is controlled by the
small subspace dimension $r$ and amortized over $K$ actor updates.
These properties help keep the additional computational and memory
costs manageable, supporting the scalability of \method\ to
large neural actors.

\section{Theoretical Analysis}
\label{sec:theory}

In this section, we theoretically analyze \method.
We first examine how the local metric in \method\ equilibrates positive and negative
curvature and establish the conditions for achieving exact curvature equilibration
(Subsection~\ref{sec:geometry-theory}). We then establish first-order
actor loss matching and gradient norm bounds
(Subsection~\ref{sec:direction-theory}), followed by an analysis
of the projected curvature comparison and its implications for
local actor loss (Subsection~\ref{sec:curvature-theory}).
Finally, we derive conditional stationarity bounds for direct
actor updates $\theta_{t+1}=\theta_t-\eta u_t$ under fixed
(Subsection~\ref{sec:fixed-theory}) and changing actor losses
(Subsection~\ref{sec:moving-theory}), and relate these results
to entropy-regularized policy optimization.
Complete proofs are provided in Appendix~\ref{app:proofs}.

In a nutshell, our theoretical results show that \method\ can equilibrate
positive and negative curvature within the search subspace under bounded
metric weights (Propositions~\ref{prop:geometry}
and~\ref{prop:feasibility}). Through first-order loss matching and gradient
mixing, \method\ can effectively reduce the actor loss,
while bounding the norm of the mixed gradient relative to the
raw gradient (Theorem~\ref{thm:matching}). Under suitable conditions, the mixed gradient
can also achieve a greater decrease in actor loss than the raw gradient
(Corollary~\ref{cor:local-comparison}).
Furthermore, we establish convergence toward first-order
stationarity for direct actor updates under standard smoothness
assumptions for fixed actor losses (Theorem~\ref{thm:stationary}),
together with conditional stationarity bounds for changing actor
losses (Theorem~\ref{thm:moving}). We finally show that convergence toward
stationarity of the expected actor loss implies approximate stationarity of
the entropy-regularized return, subject to suitable assumptions
(Proposition~\ref{prop:policy-bridge}).

\subsection{Local curvature equilibration under bounded weights}
\label{sec:geometry-theory}

We analyze the actor loss at a single update, holding the data batch,
actor noise, and auxiliary quantities fixed. For simplicity, we omit
the update index and write $\loss(\theta)$ for the resulting actor loss.
Let $Q\in\R^{p\times r}$ be the orthonormal search basis and
$f(z)=\loss(\theta+Qz)$ the actor loss restricted to the search
subspace. The projected Hessian is
$A=\nabla^2f(0)=Q^\top\nabla^2\loss(\theta)Q$.
We consider a fixed inverse metric $B\succ0$, with $G=B^{-1}$.

\begin{proposition}[Local geometry under a fixed metric]
\label{prop:geometry}
Suppose $f$ is $C^3$ near $0$. Then
\begin{align}
 \operatorname{div}_{G}(B\nabla f)(0)&=\tr(BA),
 \label{eq:divergence}\\
 \E_{v\sim\mathrm{Unif}(\mathbb S^{r-1})}
 f(\rho B^{1/2}v)
 &=f(0)+\frac{\rho^2}{2r}\tr(BA)+O(\rho^3).
 \label{eq:sphere-average}
\end{align}
For the gradient update $T_B(z)=z-\eta B\nabla f(z)$ with
sufficiently small $\eta$,
\begin{equation}
 \log|\det DT_B(0)|=-\eta\tr(BA)
 -\frac{\eta^2}{2}\tr((BA)^2)+O(\eta^3).
 \label{eq:volume}
\end{equation}
Thus, the curvature-equilibrated target $\tr(BA)=0$ eliminates
the local divergence, the quadratic term in the average actor
loss over a sphere under $G$, and the first-order local volume change.
\end{proposition}

The points $\rho B^{1/2}v$ in Equation~\eqref{eq:sphere-average}
lie on a sphere of radius $\rho$ under the local metric $G$.
These geometric properties apply to the metric-adjusted gradient
before rescaling and mixing. Whenever the rescaling factor $\chi$
varies with $\theta$, the divergence becomes
\begin{equation}
 \operatorname{div}(\chi B\nabla f)
 =\chi\tr(B\nabla^2f)+\nabla\chi^\top B\nabla f.
 \label{eq:scale-divergence}
\end{equation}
Assuming that $\chi$ is fixed, mixing with the raw gradient gives
$B_\lambda=(1-\lambda)I_r+\lambda\chi B$, for which
$\tr(B_\lambda A)=(1-\lambda)\tr(A)+\lambda\chi\tr(BA)$
need not vanish. In the full parameter space, including the
unchanged gradient component outside the search subspace,
the corresponding operator is
\begin{equation}
 P_\lambda=I_p+\lambda Q(\chi B-I_r)Q^\top\succ0,\qquad
 \tr(P_\lambda H)=\tr(H)+\lambda\tr((\chi B-I_r)Q^\top HQ),
 \label{eq:full-trace}
\end{equation}
where $H=\nabla^2\loss(\theta)$.
Therefore, the curvature-equilibrated target applies to the local
metric before rescaling and mixing, and does not generally imply
zero divergence for the final mixed gradient used for actor training.

\begin{proposition}[Exact curvature equilibration under bounded weights]
\label{prop:feasibility}
Let $\mu_i$ denote the eigenvalues of $A$, and define
$S_+=\sum_{\mu_i>0}\mu_i$ and
$S_-=\sum_{\mu_i<0}|\mu_i|$.
Exact curvature equilibration $\sum_i b_i\mu_i=0$ with
$b_i\in[m,M]$ is feasible if and only if
\begin{equation}
 mS_+-MS_-\leq0\leq MS_+-mS_-.
 \label{eq:feasibility}
\end{equation}
When $S_+,S_->0$, this is equivalent to
$S_+/S_-\in[m/M,M/m]$.
When exact curvature equilibration is feasible, we seek the
curvature-direction weights $b_i$ that deviate the least from the
Euclidean metric (i.e., $b_i=1$ for all $i$):
\begin{equation}
 \min_{b\in[m,M]^r}\tfrac12\norm{b-\mathbf1}^2
 \quad\text{subject to}\quad \mu^\top b=0.
 \label{eq:ideal-qp}
\end{equation}
This optimization problem has a unique solution
$b_i^*=\clip(1-\tau\mu_i,m,M)$, where the scalar $\tau$
is chosen to satisfy the equality constraint.
For $\mu=0$, the solution is $b^*=\mathbf1$.

For the weights $b_i$ determined according to Equation~\eqref{eq:practical-metric},
let $S=\sum_i\mu_i$ and $D=\sum_i\mu_i^2$.
The remaining weighted curvature sum is
\begin{equation}
 \sum_i b_i\mu_i=
 \frac{S\epsm}{D+\epsm}
 +\sum_i\mu_i\bigl[b_i-(1-\kappa\mu_i)\bigr].
 \label{eq:trace-residual}
\end{equation}
\end{proposition}

Equation~\eqref{eq:trace-residual} separates the effects of
regularization and clipping on curvature equilibration, motivating
the trace-reduction check in Equation~\eqref{eq:metric-eligibility}.
For a nonzero projected Hessian, exact equilibration with positive
weights requires both positive and negative curvature, together
with the bounded-weight feasibility condition in
Equation~\eqref{eq:feasibility}.

\subsection{First-order actor loss matching and gradient norm bounds}
\label{sec:direction-theory}

We analyze the mixed gradient with respect to an orthonormal search basis $Q$
and a bounded inverse metric $mI\preceq B\preceq MI$. Let $g$
denote the raw actor gradient, $a=Q^\top g$ its subspace coordinates,
and $g^\perp=(I-QQ^\top)g$ its component outside the search subspace.
The analysis also applies when the local geometry is reused between
refreshes (also see Section~\ref{sec:computational-cost}). We use $u$ to denote the proposed mixed gradient before gradient selection.

\begin{theorem}[First-order loss matching and bounded gradient norm]
\label{thm:matching}
For $a\neq0$, let
$\chi=\norm{a}^2/(a^\top Ba+\epss)$,
$d=g^\perp+Q\chi Ba$, and $u=(1-\lambda)g+\lambda d$,
where $\epss\geq0$ and $\lambda\in[0,1]$. Define
\begin{equation}
 \delta=\frac{\epss\norm{a}^2}{a^\top Ba+\epss},\qquad
 K_B=\frac{M+m}{2\sqrt{mM}},\qquad
 C_\lambda^2=1+\lambda^2(K_B^2-1).
 \label{eq:bound-constants}
\end{equation}
Then
\begin{align}
 g^\top d&=\norm{g}^2-\delta,&
 g^\top u&=\norm{g}^2-\lambda\delta
              \geq(1-\lambda)\norm{g}^2,
 \label{eq:matching-theorem}\\
 \norm{u}^2&=\norm{g}^2-2\lambda\delta
                +\lambda^2\norm{d-g}^2,&
 \norm{u}^2&\leq C_\lambda^2\norm{g}^2.
 \label{eq:norm-theorem}
\end{align}
When $\epss=0$, first-order loss matching is exact:
$g^\top u=\norm{g}^2$. Moreover, $d-g$ is orthogonal to $g$,
with $\norm{d-g}^2\leq(K_B^2-1)\norm{a}^2$.
Consequently, for $g\neq0$,
\[
 1\leq\frac{\norm{u}}{\norm{g}}\leq C_\lambda,
 \qquad \cos(g,u)\geq C_\lambda^{-1}.
\]
\end{theorem}

In our experiments, $m=1/3$, $M=3$, and $\lambda=0.3$,
giving $C_\lambda=\sqrt{1.16}\approx1.077$ and
$\cos(g,u)\geq0.928$.
These bounds show that gradient mixing limits changes in
magnitude and direction while preserving the predicted first-order
decrease in actor loss. The theoretical norm bound also falls
below the numerical acceptance limit of $1.1$ specified in
Appendix~\ref{app:acceptance-implementation}.
If gradient selection rejects the mixed gradient, the actor optimizer receives
the raw gradient $g$, which satisfies the same norm bound and
exact first-order matching.

\subsection{Projected curvature comparison and local actor loss}
\label{sec:curvature-theory}

Let $Q$ denote the local search subspace basis, constructed at
the most recent geometry refresh and reused between refreshes.
Let $\overline A$ be the corresponding stored projected Hessian
and $H=\nabla^2\loss(\theta)$ the Hessian of the current actor loss.
Let $g$ be the raw gradient and $u$ the gradient selected for
the current actor update. Define their subspace coordinates
$a=Q^\top g$ and $c=Q^\top u$, their common component outside
the search subspace $g^\perp=(I-QQ^\top)g$, and the difference
between the current and stored projected Hessians
$E=Q^\top HQ-\overline A$.

\begin{proposition}[Curvature comparison under reused local geometry]
\label{prop:cached-gap}
Let $\Delta_{\rm cache}=c^\top\overline A c-a^\top\overline A a$.
The difference between the second-order terms of the local actor loss
along \(u\) and \(g\) satisfies
\begin{align}
 \Delta_H&=u^\top Hu-g^\top Hg\\
 &=\Delta_{\rm cache}+(c-a)^\top E(c+a)
      +2(g^\perp)^\top HQ(c-a),
 \label{eq:cache-identity}\\
 \Delta_H&\leq\Delta_{\rm cache}
 +\norm{E}\norm{c-a}\norm{c+a}
 +2\norm{Q^\top Hg^\perp}\norm{c-a}.
 \label{eq:cache-bound}
\end{align}
At a refresh using the current actor loss, the stored projected
Hessian is exact (i.e., $E=0$), and the search basis contains $g$
(i.e., $g^\perp=0$). Thus, accepting a mixed gradient with lower
projected curvature guarantees $\Delta_H<0$.
Between refreshes, a sufficient condition for $\Delta_H<0$ is
\begin{equation}
 -\Delta_{\rm cache}>
 \norm{E}\norm{c-a}\norm{c+a}
 +2\norm{Q^\top Hg^\perp}\norm{c-a}.
\end{equation}
Thus, the reduction in projected curvature measured using the stored
Hessian must outweigh the effects of changes in the projected Hessian
and interactions with the gradient component outside the search subspace.
\end{proposition}

The last two error terms in Equation~\eqref{eq:cache-bound} account for
the difference $E=Q^\top HQ-\overline A$ between the current and
stored projected Hessians, and the interaction $Q^\top Hg^\perp$
between the search subspace and the gradient component outside it.
If the actor loss $\loss$ remains unchanged, its Hessian is
$L_H$-Lipschitz continuous, and
$\overline A=Q^\top\nabla^2\loss(\theta_{\rm refresh})Q$,
then
$\norm{E}\leq L_H\norm{\theta_t-\theta_{\rm refresh}}$.
During actor--critic training, changes in the critic, data batch,
and actor noise can also change the actor loss, introducing
additional differences not captured by this bound.
These effects motivate periodic refreshes of the local geometry.

\begin{corollary}[Local actor loss comparison with a raw-gradient update]
\label{cor:local-comparison}
Let $g=\nabla\loss(\theta)$ and $u=\widetilde u$ be the
accepted mixed gradient from Theorem~\ref{thm:matching}.
Assume that $\loss$ has an $L_H$-Lipschitz Hessian along
the line segments from $\theta$ to $\theta-\eta u$ and
$\theta-\eta g$, where $\eta>0$. Then
\begin{equation}
 \begin{split}
 \loss(\theta-\eta u)-\loss(\theta-\eta g)
 \leq\eta\lambda\delta+\frac{\eta^2}{2}\Delta_H
 +\frac{L_H\eta^3}{6}\bigl(\norm{u}^3+\norm{g}^3\bigr).
 \end{split}
 \label{eq:local-comparison}
\end{equation}
For a quadratic actor loss with exact first-order matching
($\epss=0$), $\Delta_H<0$ guarantees a lower loss for the
mixed-gradient update than for the raw-gradient update,
regardless of the step size $\eta>0$. For a nonquadratic loss,
the negative curvature difference must outweigh the Taylor
remainder and, when $\epss>0$, the first-order matching error
to guarantee the same result.
\end{corollary}

Corollary~\ref{cor:local-comparison} compares two direct updates from
the same actor parameters using the same actor loss. Since first-order matching
preserves the descent direction, a sufficiently small step size
ensures a decrease in actor loss. Corollary~\ref{cor:local-comparison} further
identifies conditions under which the mixed-gradient update achieves a
lower loss than the raw-gradient update.

\subsection{Convergence conditions for fixed actor losses}
\label{sec:fixed-theory}

We first analyze convergence for a fixed actor loss $\loss$ under
direct gradient updates. Let $g_t=\nabla\loss(\theta_t)$ denote
the raw actor gradient at actor update step $t$. Convergence
toward a stationary point is validated through $\norm{g_t}$.

\begin{assumption}[Conditions for direct actor updates]
\label{ass:direct}
The actor loss $\loss$ is lower bounded by $\loss_*$ and has
an $L$-Lipschitz gradient. At each actor update, the selected
gradient $u_t$ satisfies
\begin{equation}
 \ip{g_t}{u_t}\geq c_0\norm{g_t}^2,
 \qquad
 \norm{u_t}\leq C_0\norm{g_t},
 \label{eq:direct-conditions}
\end{equation}
where $c_0>0$ and $C_0\geq1$. The actor parameters are updated
by $\theta_{t+1}=\theta_t-\eta u_t$.
\end{assumption}

\begin{theorem}[Stationarity and conditional linear convergence]
\label{thm:stationary}
Under Assumption~\ref{ass:direct}, let
$0<\eta<2c_0/(LC_0^2)$ and
$a_\eta=\eta c_0-L\eta^2C_0^2/2$. Then
\begin{equation}
 \min_{0\leq t<T}\norm{g_t}^2
 \leq\frac1T\sum_{t=0}^{T-1}\norm{g_t}^2
 \leq\frac{\loss(\theta_0)-\loss_*}{T a_\eta}.
 \label{eq:stationary-rate}
\end{equation}
If $\loss$ additionally satisfies the
Polyak--\L{}ojasiewicz (PL) inequality
\[
 \tfrac12\norm{\nabla\loss(\theta)}^2
 \geq\mu_{\rm PL}(\loss(\theta)-\loss_*),
\]
then, provided $0\leq1-2\mu_{\rm PL}a_\eta<1$,
\begin{equation}
 \loss(\theta_T)-\loss_*
 \leq(1-2\mu_{\rm PL}a_\eta)^T
 \bigl(\loss(\theta_0)-\loss_*\bigr).
 \label{eq:pl-rate}
\end{equation}
\end{theorem}

Theorem~\ref{thm:matching} shows that the mixed gradient
remains aligned with the raw gradient and has a bounded norm.
With exact first-order loss matching ($\epss=0$),
Equation~\eqref{eq:direct-conditions} holds with
$c_0=1$ and $C_0=C_\lambda$.
When the denominator regularizer $\epss>0$ is introduced
in the matching scale (Equation~\eqref{eq:regularized-matching-scale}),
the norm bound remains unchanged, while
$c_0=1-\lambda>0$ provides a conservative alignment
bound for $\lambda<1$.
These bounds also hold when the raw gradient is selected
instead of the mixed gradient.
Hence, under Assumption~\ref{ass:direct},
the direct actor updates converge toward first-order
stationarity at a rate of $O(1/T)$ in the average squared
gradient norm. Linear convergence additionally requires
the PL inequality, which is not assumed to hold for general
neural actor losses.

\begin{proposition}[Accumulated improvement from accepted mixed gradients]
\label{prop:surplus}
Suppose $\loss$ satisfies Assumption~\ref{ass:direct}
and the direct actor updates additionally satisfy
\[
 \loss(\theta_{t+1})
 \leq\loss(\theta_t-\eta g_t)-I_ts_t,
\]
where $I_t\in\{0,1\}$ indicates whether the mixed gradient
is accepted, $s_t\geq0$ denotes its additional loss decrease
relative to the raw-gradient update, and $0<\eta<2/L$.
Then
\begin{equation}
 (\eta-L\eta^2/2)\sum_{t<T}\norm{g_t}^2
 +\sum_{t<T}I_ts_t
 \leq\loss(\theta_0)-\loss_*.
 \label{eq:surplus}
\end{equation}
\end{proposition}

The term $\sum_{t<T}I_ts_t$ quantifies the accumulated
additional loss decrease from accepted mixed gradients
along the training trajectory. Corollary~\ref{cor:local-comparison}
provides sufficient conditions for $s_t>0$.

\subsection{Convergence conditions for changing actor losses}
\label{sec:moving-theory}

In deep reinforcement learning, the actor loss changes as the critic
and replay data evolve. Let $\mathcal F_t$ denote the information
available before performing the current actor update.
We use $\bar{\loss}_t(\theta)$ to denote the conditional expected
actor loss under the current replay distribution and frozen critic.
Its gradient is $\bar g_t=\nabla\bar{\loss}_t(\theta_t)$, whereas
$g_t=\nabla\loss_t(\theta_t)$ is the raw gradient computed from
the sampled actor loss.

\begin{assumption}[Conditions for changing actor losses]
\label{ass:moving}
Each $\bar{\loss}_t$ has an $L$-Lipschitz gradient and shares
a common lower bound $\loss_*$. For the selected gradient $u_t$,
assume
\begin{align}
 \E[\ip{\bar g_t}{u_t}\mid\mathcal F_t]
   &\geq c_0\norm{\bar g_t}^2-\beta_t,
   \label{eq:expected-align}\\
 \E[\norm{u_t}^2\mid\mathcal F_t]
   &\leq C_0^2\norm{\bar g_t}^2+\sigma_t^2,
   \label{eq:expected-norm}\\
 \E[\bar{\loss}_{t+1}(\theta_{t+1})
     -\bar{\loss}_t(\theta_{t+1})\mid\mathcal F_t]
   &\leq v_t,
   \label{eq:objective-drift}
\end{align}
where $c_0>0$, $C_0\geq1$, and
$\beta_t,\sigma_t^2,v_t\geq0$ are integrable and
$\mathcal F_t$-measurable.
Here, $\beta_t$ bounds the expected alignment error,
$\sigma_t^2$ bounds the extra expected squared gradient norm,
and $v_t$ bounds the change in the expected actor loss.
The direct actor update is $\theta_{t+1}=\theta_t-\eta u_t$.
\end{assumption}

\begin{theorem}[Conditional stationarity for changing actor losses]
\label{thm:moving}
Under Assumption~\ref{ass:moving}, if
$a_\eta=\eta c_0-L\eta^2C_0^2/2>0$, then
\begin{equation}
 \frac1T\sum_{t<T}\E\norm{\bar g_t}^2
 \leq
 \frac{
 \E\bar{\loss}_0(\theta_0)-\loss_*+
 \sum_{t<T}\E[v_t+\eta\beta_t+(L\eta^2/2)\sigma_t^2]
 }{T a_\eta}.
 \label{eq:moving-rate}
\end{equation}
For a fixed expected actor loss with $\beta_t=v_t=0$
and uniformly bounded $\sigma_t^2$, choose $\eta$
proportional to $T^{-1/2}$.
This gives an $O(T^{-1/2})$ stationarity bound.
Persistent alignment errors and variations of the
expected actor loss can weaken the convergence
guarantee toward stationarity.
\end{theorem}

First-order loss matching ensures alignment of the mixed gradient
with the sampled raw gradient $g_t$, whereas
Equation~\eqref{eq:expected-align} requires alignment with
the expected gradient $\bar g_t$.
To examine this distinction, suppose $g_t$ is conditionally
unbiased and write $u_t=g_t+h_t$, where $h_t$ captures the
gradient adjustment. Then
\begin{equation}
 \bar g_t^\top\E[u_t\mid\mathcal F_t]
 =\norm{\bar g_t}^2+
 \E[(\bar g_t-g_t)^\top h_t\mid\mathcal F_t]
 +\E[g_t^\top h_t\mid\mathcal F_t].
 \label{eq:noise-correlation}
\end{equation}
Exact first-order matching eliminates the last term, but the
gradient adjustment $h_t$ may still be correlated with the
sampling error $g_t-\bar g_t$. Reusing the same batch for
critic and actor updates can also affect the unbiasedness of
$g_t$. Therefore, first-order matching with $g_t$ does not
guarantee the expected-gradient alignment required
by Equation~\eqref{eq:expected-align}.

We can bound this effect without assuming independence.
Suppose $g_t$ is conditionally unbiased, the gradient adjustment
$h_t=u_t-g_t$ satisfies exact first-order matching
$g_t^\top h_t=0$ for every sampled actor gradient, and
\[
 \E[\norm{g_t-\bar g_t}^2\mid\mathcal F_t]
 \leq\sigma_{g,t}^2.
\]
By Theorem~\ref{thm:matching},
$\norm{h_t}\leq\kappa_h\norm{g_t}$, where
$\kappa_h=\lambda\sqrt{K_B^2-1}$.
Then, for any $\epsilon\in(0,1)$,
\begin{equation}
 \E[\bar g_t^\top u_t\mid\mathcal F_t]
 \geq(1-\epsilon)\norm{\bar g_t}^2
 -\left(\kappa_h+\frac{\kappa_h^2}{4\epsilon}\right)
 \sigma_{g,t}^2,
 \label{eq:conservative-bias}
\end{equation}
and
\[
 \E[\norm{u_t}^2\mid\mathcal F_t]
 \leq C_\lambda^2(\norm{\bar g_t}^2+\sigma_{g,t}^2).
\]
These bounds quantify how sampling variability affects
gradient alignment and norm in Assumption~\ref{ass:moving}.

\begin{proposition}[Relation to the entropy-regularized return]
\label{prop:policy-bridge}
Assume a differentiable action reparameterization
$a_\theta(s,\xi)$ and the regularity conditions of the
soft policy-gradient theorem. The distribution of $\xi$
is independent of $\theta$. For each state, assume that
the sampled actor loss is differentiable with respect
to $\theta$ almost surely when the critic is held fixed.
Assume also that differentiation can be interchanged
with expectation over $\xi$.
Using the exact critic $Q_\alpha^{\pi_\theta}$ and the
policy's discounted state distribution $d_\theta$,
the actor gradient satisfies
\begin{equation}
 g_\theta^*=\E_{s\sim d_\theta,\xi}
 \nabla_\theta\bigl[
 \alpha\log\pi_\theta(a_\theta(s,\xi)\mid s)
 -Q_\alpha^{\pi_\theta}(s,a_\theta(s,\xi))
 \bigr]_{Q_\mathrm{fixed}}
 =-(1-\gamma)\nabla J_\alpha(\theta),
 \label{eq:soft-pg-bridge}
\end{equation}
where $J_\alpha(\theta)$ is the expected entropy-regularized
return defined in Equation~\eqref{eq:soft-objective}.
Here, differentiation passes through the sampled action
while treating the critic as fixed with respect to the
actor parameters.

In contrast, $\bar g_t$ is the expected actor gradient
computed using the replay state distribution and a learned
critic. We define its difference from the exact actor
gradient as
\[
 e_t=\norm{\bar g_t-g_{\theta_t}^*}
 =\norm{\bar g_t+(1-\gamma)\nabla J_\alpha(\theta_t)}.
\]
Then
\begin{equation}
 \frac1T\sum_{t<T}\E [ \norm{\nabla J_\alpha(\theta_t)}^2 ]
 \leq\frac{2}{(1-\gamma)^2}
 \left[
 \frac1T\sum_{t<T}\E [ \norm{\bar g_t}^2 ]
 +\frac1T\sum_{t<T}\E [ e_t^2 ]
 \right].
 \label{eq:return-stationarity}
\end{equation}

Furthermore, suppose the exact state-wise actor gradient
has norm at most $G_\psi$, the action Jacobian satisfies
$\norm{D_\theta a_\theta}\leq A_\theta$, and the learned
critic's action-gradient error is uniformly bounded by
$\epsilon_{Q,t}$. Then
\begin{equation}
 e_t\leq
 2G_\psi\TV(\nu_t^{\rm replay},d_{\theta_t})
 +A_\theta\epsilon_{Q,t},
 \label{eq:replay-critic-bias}
\end{equation}
where $\nu_t^{\rm replay}$ denotes the replay state
distribution and $\TV$ is the total variation distance.
\end{proposition}
Equation~\eqref{eq:soft-pg-bridge} follows the soft
policy-gradient theorem~\citep{shi2019softpg}.
Equation~\eqref{eq:replay-critic-bias} bounds the gradient
error arising from replay distribution mismatch and critic's
action-gradient error. Since SAC uses the minimum of two
learned critics, this bound holds whenever the minimum is
differentiable and both critics satisfy the assumed error bound.

Together, Theorem~\ref{thm:moving} and
Proposition~\ref{prop:policy-bridge} provide a conditional
stationarity bound for the entropy-regularized return,
provided that gradient alignment errors, sampling variability,
critic errors, replay distribution mismatch, and changes
in the expected actor loss can be properly controlled.

\section{EQUIP-SAC: Integration with Soft Actor--Critic}
\label{sec:sac}

We integrate \method\ into Soft Actor--Critic (SAC)
\citep{haarnoja2018sac,haarnoja2018sacapplications},
resulting in \sacmethod. The integration modifies only the
gradient supplied to the actor optimizer, leaving the critic
learning, replay sampling, and environment interaction
procedures unchanged.

\subsection{SAC actor and critic optimization}

\sacmethod\ follows the learning framework of SAC, with a stochastic
actor $\pi_\theta$, two learned action-value functions
$Q_{\omega_1}$ and $Q_{\omega_2}$, their target networks
$Q_{\overline\omega_1}$ and $Q_{\overline\omega_2}$,
and a replay buffer $\mathcal D$. The entropy temperature
$\alpha>0$ is fixed.

The actor uses a squashed Gaussian policy with differentiable
action sampling:
\begin{equation}
 \begin{aligned}
 x_\theta(s,\xi)&=\mu_\theta(s)+\sigma_\theta(s)\odot\xi,
 \quad \xi\sim\mathcal N(0,I),\\
 a_\theta(s,\xi)&=c_{\rm act}\tanh (x_\theta(s,\xi)),
 \end{aligned}
 \label{eq:squashed-policy}
\end{equation}
where $\mu_\theta$ and $\sigma_\theta$ are the actor's
predicted mean and standard deviation, and $c_{\rm act}$
is the action scaling factor~\footnote{
Following the Spinning Up SAC implementation, $c_{\rm act}$
is set to the common symmetric action bound of each environment.
It is $1$ for HalfCheetah, Walker2d, Ant, and Swimmer,
$0.4$ for Humanoid, and $2$ for Pusher.}.
Since the $\tanh$ transformation changes the
probability density of the Gaussian actions,
the action log-probability is computed from the
original Gaussian density with
a Jacobian correction~\citep{haarnoja2018sac}.

For a minibatch
$\mathcal B_t=\{(s_i,a_i,r_i,s_i')\}_{i=1}^N$
sampled from $\mathcal D$, the critic target is
\begin{equation}
 y_i=r_i+\gamma
 \left[
 \min_{j=1,2}Q_{\overline\omega_j}(s_i',a_i')
 -\alpha\log\pi_{\theta_t}(a_i'\mid s_i')
 \right],
 \label{eq:sac-target}
\end{equation}
where $a_i'=a_{\theta_t}(s_i',\xi_i')$ is sampled from the
current actor. The two critics minimize the sum of their
mean squared Bellman errors:
\begin{equation}
 L_Q(\omega_1,\omega_2)=\frac1N\sum_{i=1}^N
 \sum_{j=1}^2
 \left(Q_{\omega_j}(s_i,a_i)-y_i\right)^2.
 \label{eq:critic-loss}
\end{equation}
The target critics are updated using Polyak averaging:
\begin{equation}
 \overline\omega_j
 \leftarrow\rho\overline\omega_j+(1-\rho)\omega_j,
 \qquad j\in\{1,2\},
 \label{eq:polyak}
\end{equation}
where $\rho\in(0,1)$ controls the update rate.
After updating the critics, their parameters are held fixed
during actor optimization. Using the states from the same
minibatch and newly sampled actor noise, the SAC actor loss is
\begin{equation}
 \loss_t^{\rm SAC}(\theta)=\frac1N\sum_{i=1}^N
 \left[
 \alpha\log\pi_\theta(a_\theta(s_i,\xi_i)\mid s_i)
 -\min_{j=1,2}Q_{\omega_j}(s_i,a_\theta(s_i,\xi_i))
 \right].
 \label{eq:sac-actor}
\end{equation}
This loss approximates the soft policy improvement objective
in Equation~\eqref{eq:gibbs-kl}. The reparameterized actions
allow actor loss gradients and HVPs to be computed with respect
to the actor parameters while keeping the critic parameters fixed.

\subsection{Integration of \method\ into SAC actor updates}

At actor update $t$, \method\ takes the SAC actor loss
$\loss_t^{\rm SAC}$ in Equation~\eqref{eq:sac-actor} as input
and computes the raw gradient
$g_t=\nabla_\theta\loss_t^{\rm SAC}(\theta_t)$.
Algorithm~\ref{alg:mechanism} then uses the stored local
geometry to construct and select the gradient $u_t$,
which is supplied to the actor's Adam optimizer.

Whenever the local geometry is refreshed, \method\ constructs
the search subspace from the current gradient and gradient history,
and computes the projected Hessian using HVPs
(see Section~\ref{sec:method}).
All HVPs are computed using the same replay minibatch, sampled actor noise,
and frozen critics as the raw gradient, ensuring that they
correspond to the same actor loss. Between refreshes, the
stored local geometry is reused, and only the current
actor gradient is computed. After each successful actor
update, the selected gradient $u_t$ is added to the
gradient history.
Algorithm~\ref{alg:sac} summarizes the complete
training procedure.

\begin{algorithm}[t]
\caption{\sacmethod: SAC with curvature-equilibrated actor optimization}
\label{alg:sac}
\begin{algorithmic}[1]
\Require Actor $\theta$, critics $\omega_1,\omega_2$,
 target critics, replay buffer $\mathcal D$,
 fixed temperature $\alpha$, Adam optimizers,
 and \method\ state
\For{each environment interaction $k$}
 \State Sample an action using the initial random-exploration
 policy or the current actor $\pi_\theta$
 \State Execute the action and store the transition in $\mathcal D$
 \If{an update burst is scheduled after warmup}
  \For{each actor--critic update}
   \State Sample a minibatch $\mathcal B_t$ from $\mathcal D$
   \State Compute critic targets using Equation~\eqref{eq:sac-target}
   \State Update both critics using Equation~\eqref{eq:critic-loss}
   \State Freeze the updated critics and sample actor noise $\xi$
   \State Evaluate the actor loss $\loss_t^{\rm SAC}$
   using Equation~\eqref{eq:sac-actor}
   \State Apply Algorithm~\ref{alg:mechanism} to update the actor
   using Adam with the selected gradient $u_t$
   \State Store $u_t$ in the gradient history
   \State Update target critics using Equation~\eqref{eq:polyak}
  \EndFor
 \EndIf
\EndFor
\end{algorithmic}
\end{algorithm}

As evidenced in Algorithm~\ref{alg:sac}, \method\
adjusts the gradient supplied to the actor optimizer
while preserving the original SAC actor loss and critic learning
procedure. This approach also makes \method\ applicable to other
gradient-based actor--critic algorithms, provided that their actor losses
support the required gradient and curvature computations.

Detailed implementation settings
are presented in Appendix~\ref{app:settings}, while the configurations
used in our experiments are reported in
Section~\ref{sec:experiments}.

\section{Experiments}
\label{sec:experiments}

In this section, we conduct experiments to evaluate
the sample efficiency and learning performance of
\sacmethod\ over 300,000 environment interactions.
We compare \sacmethod\ with SAC \citep{haarnoja2018sac},
an adapted projected saddle-free Newton (PN) baseline
following \citet{dauphin2014identifying}, and VR-SCP
\citep{khorasani2025vrscp}, a recent second-order
policy optimization method.
We also conduct ablation studies to investigate the effectiveness
of first-order loss matching, trajectory-informed search
subspaces, and curvature-based gradient selection mechanisms.
Finally, we examine the impact of curvature estimation
effort, parameter sensitivity, and gradient adjustment
acceptance on learning performance.

\subsection{Experimental setup}
\label{sec:exp-setup}

\paragraph{Benchmark problems and SAC-based methods.}
We conduct experiments on 6 MuJoCo-v5 continuous control benchmarks, including
HalfCheetah, Walker2d, Ant, Humanoid, Swimmer, and Pusher~\citep{todorov2012mujoco}.
SAC, \sacmethod, and projected saddle-free Newton (PN) share
the same PyTorch SAC implementation, adapted from OpenAI
Spinning Up~\citep{achiam2018spinningup}\footnote{Implementation
details and default hyper-parameters are available at
\url{https://spinningup.openai.com/en/latest/algorithms/sac.html}.}.

We also follow the default hyper-parameter settings provided by OpenAI
Spinning Up. Specifically, the actor and twin critics
each use two 256-unit ReLU hidden layers, with a tanh-squashed
Gaussian policy for the actor. 
We use Adam with a learning rate of $10^{-3}$ for both
actor and critic optimization, a discount factor of
$\gamma=0.99$, a fixed entropy temperature of $\alpha=0.2$,
and Polyak retention of $0.995$.
Actor and critic updates use minibatches of 100 transitions
randomly sampled from the replay buffer.
Actor--critic training begins after 1,000 environment steps,
with 50 gradient updates performed after every 50 steps of
environmental interactions.
Furthermore, SAC, \sacmethod, and PN takes random actions during the
first 10,000 environmental steps before switching to the learned policy
for subsequent environmental explorations.

\paragraph{\method\ configuration.}
While SAC passes the raw actor gradient to Adam, \sacmethod\
adjusts it using Algorithm~\ref{alg:mechanism}.
We follow a gradient mixing weight of $\lambda=0.3$ and request
an adjustment at every actor update.
Only adjustments satisfying the metric eligibility conditions
in Subsection~\ref{sec:curvature-balancing}
(Equation~\eqref{eq:metric-eligibility}) and the gradient-selection
criterion in Subsection~\ref{sec:gradient-selection}
(Equation~\eqref{eq:curvature-gate}) are accepted.
Otherwise, the raw gradient is used.

\paragraph{VR-SCP baseline and implementation.}
We experimentally compare \sacmethod\ with VR-SCP, a recent
second-order policy optimization method proposed by
\citet{khorasani2025vrscp}.
Unlike PN, VR-SCP is an independent policy optimization algorithm
rather than an adaptation of SAC.
This comparison allows us to evaluate \sacmethod\ against a
recent method that also exploits curvature information to improve
policy optimization and sample efficiency, but adopts a different
design strategy. We adapt the authors' released implementation
to our PyTorch/Gymnasium MuJoCo-v5 environment~\citep{paszke2019pytorch,towers2025gymnasium}
while preserving its optimization procedure and key algorithmic components.

We use the hyperparameter settings in the authors'
released code\footnote{The original VR-SCP implementation
is available at \url{https://github.com/sadegh16/VR-SCP}.}, as summarized in
Table~\ref{tab:exp-vr-settings}.
All training interactions are counted toward the same
environment-step budget as \sacmethod. We have verified the numerical behavior
of all key components against the original implementation.
Thus, our baseline faithfully follows the proposed VR-SCP
algorithm, with adaptations limited to the runtime and
evaluation settings to ensure a fair comparison with
\sacmethod.

\begin{table}[t]
\caption{VR-SCP hyper-parameters for the 6 benchmark tasks.}
\label{tab:exp-vr-settings}
\centering\small
\begin{tabular}{@{}lccccc@{}}
\toprule
Task & \texttt{l} & \texttt{ro} & \texttt{S}
     & Inner iterations & KL limit\\
\midrule
Walker2d & 50 & 50 & 2 & 250 & 0.01\\
Humanoid & 400 & 1 & 5 & 1000 & 0.10\\
Other four tasks & 100 & 1 & 3 & 100 & 0.10\\
\bottomrule
\end{tabular}
\vspace{2pt}

{\footnotesize\emph{Note.} The other four tasks are HalfCheetah,
Ant, Swimmer, and Pusher. Parameter \texttt{ro} controls
the cubic regularization penalty.}
\end{table}

\paragraph{Projected saddle-free Newton baseline.}
The \emph{projected saddle-free Newton} (PN) baseline
is an adapted version of the saddle-free Newton algorithm~\citet{dauphin2014identifying}
for second-order actor optimization in SAC.
PN follows the same trajectory-informed search subspace construction,
first-order loss matching, gradient mixing and gradient selection mechanisms
as \sacmethod. The main difference lies in
how the projected Hessian is used to adjust the actor gradient.

While \sacmethod\ seeks to equilibrate positive and negative curvature,
PN aims to reduce the
influence of those directions with large curvature magnitudes.
Given the eigenvalues $\{\mu_{t,i}\}_{i=1}^{r_t}$ of the projected
actor-loss Hessian, PN computes
\begin{equation}
\begin{gathered}
s_t^{\rm PN}=\max\!\left(
\sqrt{\frac{1}{r_t}\sum_i\mu_{t,i}^2},10^{-8}\right),
\qquad
\delta_t^{\rm PN}=10^{-8}+0.1s_t^{\rm PN},\\
b_{t,i}^{\rm PN}=\clip\!\left(
\frac{s_t^{\rm PN}}{|\mu_{t,i}|+\delta_t^{\rm PN}},m,M\right).
\end{gathered}
\label{eq:exp-pn-weights}
\end{equation}
Here, $s_t^{\rm PN}$ normalizes the curvature scale,
$\delta_t^{\rm PN}$ prevents excessively large weights near
zero eigenvalues, and the bounds $[m,M]$ limit gradient rescaling.
The computed weights in Equation~\eqref{eq:exp-pn-weights} define the
inverse metric $B_t^{\rm PN}=U_t\diag(b_t^{\rm PN})U_t^\top$,
which rescales the actor gradient along the eigenvector directions
of the projected Hessian.

In our experiments, PN uses the same metric eligibility
checks as \sacmethod. Similar to \sacmethod, we set the maximum search subspace
rank $r_{\max}=32$ and history capacity $h=32$ for all 6 benchmark tasks.
Meanwhile, the refresh interval $K=64$. Both methods follow the same subspace
construction procedure, although their actual subspaces
may differ as training progresses.

\paragraph{Evaluation settings.}

We evaluate each learned policy every 4,000 environment interactions
using 10 deterministic episodes. All results are reported over the
first 300,000 environment interactions.
Our primary measure of sample efficiency is the \emph{area under
the evaluation-return curve} (AUC), averaged over the
300,000-step training window. We also compare the
\emph{evaluation returns} achieved by all competing algorithms
at 300,000 environment steps.

Each algorithm uses 10 training seeds per benchmark task.
For paired comparisons, we use matching seed IDs.
Evaluation returns are averaged across episodes within each
training seed. We report the mean performance together with
95\% Student-$t$ confidence intervals across training seeds.
The $\pm$ values in the experimental result tables
represent confidence intervals, rather than standard deviations.

For learning curves, we apply a five-checkpoint moving average
to each seed after truncating the data to 300,000 steps.
The curves show mean evaluation returns with point-wise
95\% confidence intervals. All reported AUC and final-return
values are computed from raw performance data obtained through
our experiments.

\subsection{Main benchmark results}
\label{sec:exp-main}

\begin{table}[t]
\caption{Evaluation-return AUC over the first 300k environment
interactions. Higher values indicate better sample efficiency.
Entries are means $\pm$ 95\% Student-$t$
confidence intervals.}
\label{tab:exp-main-auc}
\centering
\small
\setlength{\tabcolsep}{3pt}
\begin{tabular}{@{}lrrrr@{}}
\toprule
Task & SAC & PN (adapted SFN) & VR-SCP & \sacmethod\\
\midrule
HalfCheetah & $6434.03\pm157.50$ & $6485.50\pm194.87$ & $-587.09\pm86.30$ & $6766.84\pm219.91$\\
Walker2d & $1151.50\pm97.85$ & $1289.20\pm140.92$ & $80.21\pm19.48$ & $1537.51\pm137.58$\\
Ant & $949.70\pm66.72$ & $949.27\pm49.07$ & $-85.25\pm6.52$ & $1003.32\pm60.11$\\
Humanoid & $641.13\pm32.30$ & $639.43\pm24.07$ & $113.11\pm8.59$ & $693.05\pm37.03$\\
Swimmer & $38.54\pm0.38$ & $38.26\pm0.42$ & $0.26\pm5.94$ & $39.09\pm0.63$\\
Pusher & $-30.73\pm0.68$ & $-31.45\pm0.55$ & $-87.21\pm5.98$ & $-30.13\pm0.43$\\
\bottomrule
\end{tabular}
\end{table}

\begin{table}[t]
\caption{Evaluation returns at 300k environment interactions. Higher is better.
Entries are means $\pm$ 95\% Student-$t$ confidence intervals.}
\label{tab:exp-endpoint}
\centering
\small
\setlength{\tabcolsep}{3pt}
\begin{tabular}{@{}lrrrr@{}}
\toprule
Task & SAC & PN (adapted SFN) & VR-SCP & \sacmethod\\
\midrule
HalfCheetah & $8874.63\pm349.27$ & $8924.06\pm379.16$ & $-576.05\pm146.34$ & $9425.10\pm377.90$\\
Walker2d & $2219.67\pm435.98$ & $2922.54\pm312.80$ & $205.91\pm37.97$ & $3191.98\pm428.85$\\
Ant & $1858.24\pm270.17$ & $1909.79\pm173.76$ & $-54.10\pm13.78$ & $2232.58\pm293.21$\\
Humanoid & $783.11\pm42.34$ & $1003.72\pm113.55$ & $121.53\pm10.47$ & $1312.27\pm305.57$\\
Swimmer & $40.65\pm0.33$ & $40.16\pm0.36$ & $-1.55\pm8.51$ & $41.25\pm0.39$\\
Pusher & $-25.76\pm0.85$ & $-28.07\pm1.68$ & $-117.74\pm13.57$ & $-25.41\pm0.44$\\
\bottomrule
\end{tabular}
\end{table}

\begin{figure}[t]
\centering
\IfFileExists{vr_scp_comparison_300k.pdf}{%
 \includegraphics[width=\linewidth]{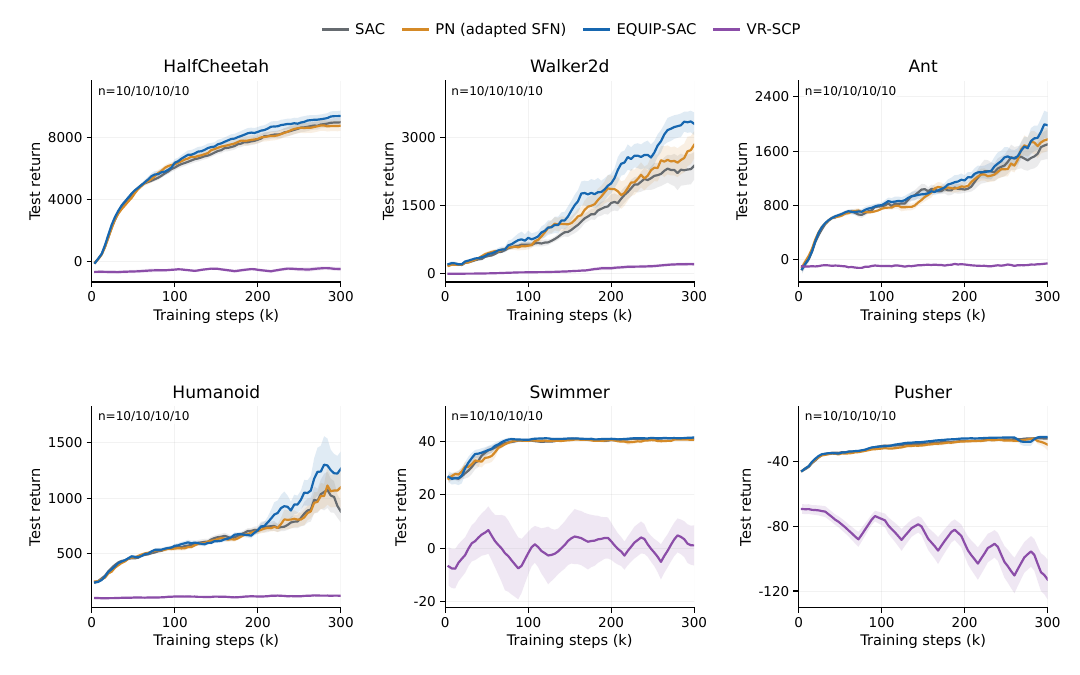}%
}{%
 \fbox{\parbox[c][1.10in][c]{0.93\linewidth}{\centering
 Copy \texttt{vr\_scp\_comparison\_300k.pdf} into the
 LaTeX project to display the four-method learning curves.}}%
}
\caption{Evaluation-return curves for SAC, PN (adapted SFN),
VR-SCP, and \sacmethod\ on 6 MuJoCo-v5 tasks over the first
300k environment interactions. Curves are smoothed separately
for each training seed using a centered five-checkpoint moving
average. Lines show mean returns across seeds, and shaded bands
indicate point-wise 95\% Student-$t$ confidence intervals.
Each checkpoint averages 10 deterministic evaluation episodes.
All methods use 10 training seeds.}
\label{fig:exp-learning-curves}
\end{figure}

\paragraph{Comparison with SAC and PN.}
Table~\ref{tab:exp-main-auc} shows that \sacmethod\ achieves the
highest mean AUC among the four competing methods on all 6 benchmark tasks.
The largest relative gain over SAC occurs on Walker2d, where
mean AUC increases from $1151.50$ to $1537.51$, an improvement
of $33.52\%$. HalfCheetah, Ant, and Humanoid achieve
smaller mean AUC gains over SAC of approximately $5.2\%$,
$5.65\%$, and $8.1\%$, respectively. Swimmer and Pusher show
more modest improvements of about $1.4\%$ and $1.9\%$.

The evaluation returns in Table~\ref{tab:exp-endpoint} show a
similar pattern, with \sacmethod\ achieving the highest mean
return at 300k on all 6 benchmark tasks. The improvements are particularly
pronounced on Walker2d, Ant, and Humanoid. Compared with SAC,
the mean return increases from $2219.67$ to $3191.98$ on Walker2d,
from $1858.24$ to $2232.58$ on Ant, and from $783.11$ to $1312.27$
on Humanoid. These results also exceed those of PN, which achieves
$2922.54$, $1909.79$, and $1003.72$ on the respective tasks.
HalfCheetah shows a smaller improvement over SAC, from $8874.63$
to $9425.10$. The gains on Swimmer and Pusher are more modest,
although \sacmethod\ still achieves the highest mean return
on both tasks.

The key difference between \sacmethod\ and PN lies in their
spectral weighting strategies. While \sacmethod\ equilibrates
positive and negative curvature, PN uses inverse-absolute-curvature
weighting following the saddle-free Newton approach.
As shown in Tables~\ref{tab:exp-main-auc} and~\ref{tab:exp-endpoint},
\sacmethod\ achieves higher mean AUC and evaluation returns
than PN on majority of the benchmark tasks.
These results demonstrate that curvature-equilibration
weighting can outperform Newton-inspired spectral weighting
for actor optimization.

\paragraph{Comparison with VR-SCP.}
Compared with SAC and PN, \sacmethod\ demonstrates a substantially
larger performance advantage over VR-SCP. It achieves higher mean
AUC and evaluation returns at 300k on all 6 benchmark tasks.
The matched-seed AUC analysis, covering all 60 comparisons,
confirms that \sacmethod\ significantly outperforms VR-SCP
at the 0.05 significance level, providing strong evidence
of its sample efficiency under the same environment-interaction budget.
However, this is an end-to-end comparison of different policy
optimization algorithms. VR-SCP uses a different training objective,
sampling scheme, and update procedure.
We follow the hyper-parameter settings provided by
\citet{khorasani2025vrscp}, using the recommended settings
without performing any additional hyper-parameter fine-tuning.

\paragraph{Learning curves.}
Figure~\ref{fig:exp-learning-curves} illustrates how the
performance differences develop throughout training.
On HalfCheetah, \sacmethod\ gradually gains an advantage
over SAC and PN as training progresses.
Walker2d shows the clearest improvement, with \sacmethod\
learning faster and achieving increasingly higher returns
during the second half of training.
On Ant and Humanoid, the three SAC-based methods perform
similarly during much of early training, but \sacmethod\
achieves higher mean returns toward 300k steps.
On Swimmer, all three methods quickly reach similar returns,
while their performance differences on Pusher remain small.
In contrast, VR-SCP consistently achieves lower returns
than the SAC-based methods across all 6 benchmark tasks.
Overall, these curves demonstrate the effectiveness
of \sacmethod.

\subsection{Controlled comparisons and ablations}
\label{sec:exp-controlled}

We investigate three key components of \method\
through ablation studies on HalfCheetah and Walker2d,
comparing the full \sacmethod\ with three variants.
The first variant removes first-order loss matching by setting
the matching scale $\chi_t=1$ and disabling the corresponding
matching checks. The second variant removes curvature-based gradient
selection while retaining the other safety checks.
The third variant replaces historical actor-gradient directions
with random directions while preserving the current raw
gradient in the search subspace. All variants follow the
same hyper-parameter settings as the main experiments of \sacmethod.

\begin{table}[t]
\caption{Component ablations of \sacmethod\ over
300k environment interactions. Entries are mean AUC
$\pm$ 95\% Student-$t$ confidence intervals.}
\label{tab:exp-component-ablation}
\centering
\small
\setlength{\tabcolsep}{4pt}
\begin{tabular}{@{}lcc@{}}
\toprule
Variant & HalfCheetah & Walker2d\\
\midrule
Full \sacmethod
 & $6766.8\pm219.9$ & $1537.5\pm137.6$\\
Without first-order matching
 & $6384.7\pm269.0$ & $1044.4\pm54.0$\\
Without curvature selection
 & $6487.9\pm227.2$ & $1443.7\pm121.9$\\
Random historical gradients
 & $6032.7\pm189.7$ & $1491.4\pm169.4$\\
\bottomrule
\end{tabular}
\end{table}

\begin{wrapfigure}{r}{0.67\textwidth}
\centering
\includegraphics[width=\linewidth]
{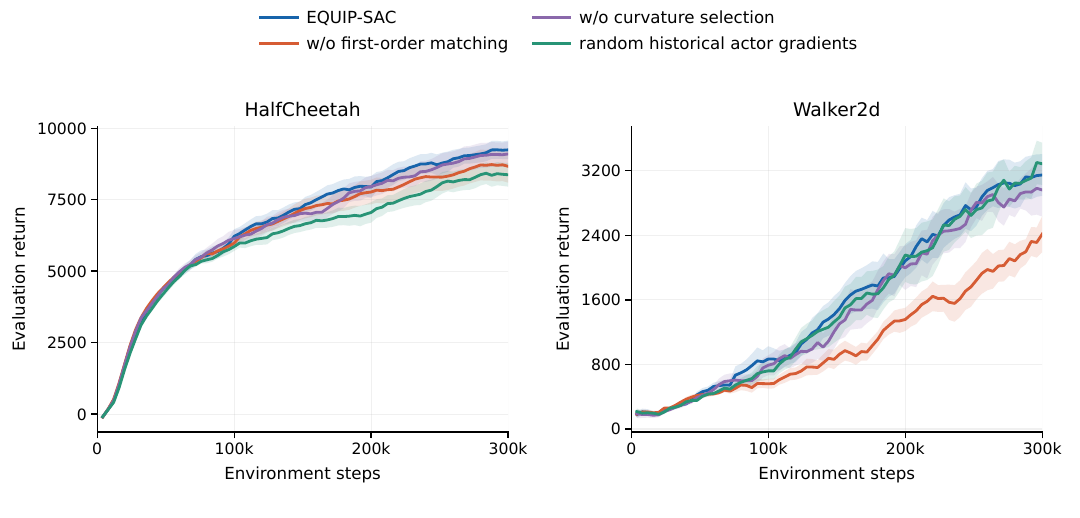}
\caption{Learning curves of \sacmethod\ and its component
ablations on HalfCheetah and Walker2d over
300k environment interactions. Lines show mean evaluation
returns with shaded 95\% Student-$t$ confidence intervals.
A five-checkpoint moving average is applied within each
seed for visualization.}
\label{fig:exp-component-ablation}
\end{wrapfigure}

Table~\ref{tab:exp-component-ablation} and
Figure~\ref{fig:exp-component-ablation} present the
ablation results. The full \sacmethod\ achieves the
highest mean AUC on both HalfCheetah and Walker2d.

\paragraph{First-order loss matching.}
Removing first-order loss matching substantially reduces
performance on Walker2d. The mean AUC decreases from
$1537.5$ to $1044.4$, a reduction of $32.1\%$.
The paired AUC difference is $493.1$, with a 95\%
confidence interval of $[317.9,668.3]$.
The effect also extends to the later stages of
training, where the mean return over the last five
evaluation checkpoints decreases from $3191.7$ to $2324.1$.
On HalfCheetah, removing first-order loss matching also
reduces mean AUC, but at a more moderate scale.
These results highlight the importance of first-order
loss matching.

\paragraph{Trajectory-informed search subspace.}
We examine the contribution of historical actor gradients
by replacing them with random directions while retaining
the current raw gradient.
On HalfCheetah, this reduces mean AUC from $6766.8$
to $6032.7$, a decrease of $10.85\%$.
The paired difference is $734.1$, with a 95\%
confidence interval of $[484.0,984.3]$.
The mean return over the last five checkpoints also
decreases from $9425.8$ to $8408.2$.
On Walker2d, however, the difference in mean AUC is much smaller.
While the full \sacmethod\ achieves a slightly higher AUC,
the random-history variant obtains a slightly higher mean
final evaluation return. These results suggest that
trajectory-informed search subspaces can improve actor
optimization, but their benefits may be task-dependent.

\paragraph{Curvature-based gradient selection.}
We evaluate the contribution of curvature-based gradient selection
by removing the criterion in Equation~\eqref{eq:curvature-gate},
while retaining all other safety checks. Given the approximately
matched first-order decrease in actor loss, this criterion favors the
mixed gradient when the projected quadratic model predicts
a lower actor loss than the raw gradient.
Removing this criterion reduces mean AUC from
$6766.8$ to $6487.9$ on HalfCheetah and from $1537.5$
to $1443.7$ on Walker2d. These results suggest that curvature-based gradient
selection can potentially improve actor optimization, with no observed
reduction in mean AUC in our experiments.

\paragraph{Curvature estimation effort.}
The projected Hessian is estimated from sampled actor losses,
which may introduce substantial estimation variance.
Averaging projected Hessians across multiple minibatches
may reduce this variance and potentially improve the reliability
of curvature-based gradient adjustments.
We investigate whether increasing the curvature estimation effort (i.e., the
number of minibatches used for computing the projected Hessian) may
allow \sacmethod\ to perform better.
Let $n_{\mathrm{curv}}$ denote the number of curvature minibatches.
We compare the standard single-minibatch estimation
($n_{\mathrm{curv}}=1$) with four-minibatch averaging
($n_{\mathrm{curv}}=4$), using the same search subspace.
For each benchmark task, both settings follow identical configurations
adopted in our main experiments reported in Section~\ref{sec:exp-main}.

Table~\ref{tab:exp-m4-anchor} shows that averaging projected
Hessians across four minibatches reduces mean AUC on five
of the 6 benchmark tasks. The largest decrease occurs on
HalfCheetah, while Walker2d is the only task showing a small
improvement. Overall, these results indicate that averaging
multiple projected Hessians does not offer substantial performance
benefit, supporting the use of the simpler and more efficient
single-minibatch estimation in \sacmethod.

\begin{table}[t]
\caption{Effect of averaging projected Hessians across four
minibatches. Values show the mean AUC differences
($n_{\mathrm{curv}}=4$ minus $n_{\mathrm{curv}}=1$),
with 95\% Student-$t$ confidence intervals.}
\label{tab:exp-m4-anchor}
\centering
\small
\begin{tabular}{@{}lrrc@{}}
\toprule
Task & Mean $\Delta$AUC & 95\% CI & Positive/negative\\
\midrule
HalfCheetah & $-584.5$ & $[-873.1,-296.0]$ & $3/7$\\
Walker2d & $26.3$ & $[-154.4,206.9]$ & $6/4$\\
Ant & $-74.4$ & $[-144.0,-4.8]$ & $3/7$\\
Humanoid & $-94.4$ & $[-133.1,-55.8]$ & $2/8$\\
Swimmer & $-0.397$ & $[-0.753,-0.040]$ & $3/7$\\
Pusher & $-1.652$ & $[-2.075,-1.229]$ & $1/9$\\
\bottomrule
\end{tabular}
\end{table}

\begin{wrapfigure}{r}{0.67\textwidth}
\centering
\includegraphics[width=0.95\linewidth]
{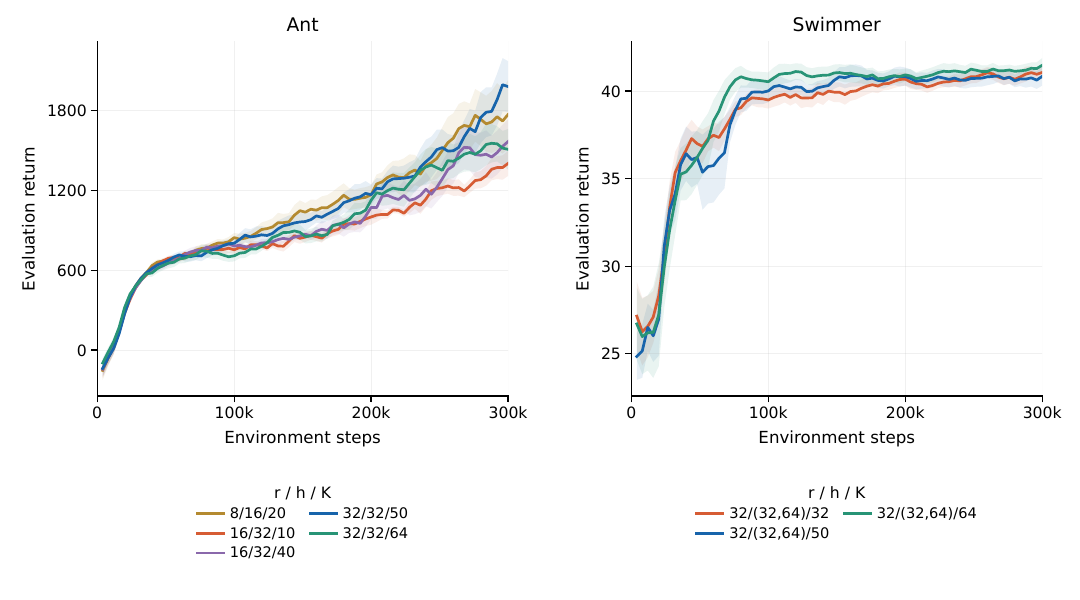}
\caption{Learning curves for \sacmethod\ with different search subspace and
refresh settings on Ant and Swimmer over 300k environment interactions.
Legends denote $r_{\max}/h/K$. Lines show mean evaluation returns
across 10 seeds, with shaded bands indicating point-wise 95\%
Student-$t$ confidence intervals. Curves are smoothed using a
five-checkpoint moving average per seed. On Swimmer, $h=32$ and
$h=64$ yield identical curves and are shown together for each $K$.}
\label{fig:exp-geometry-sensitivity}
\end{wrapfigure}

\subsection{Parameter sensitivity}
\label{sec:exp-parameters}

We investigate how the search subspace and geometry refresh
settings affect the performance of \sacmethod\ on Ant and Swimmer.
The maximum subspace rank $r_{\max}$ limits the number of
independent directions used to construct the search subspace,
within which the projected Hessian of actor loss is estimated.
The history capacity $h$ determines how many
gradients from previous actor updates are stored.
The search subspace is constructed from the current raw gradient
and up to $r_{\max}-1$ recent historical gradients.
The refresh interval $K$ determines how often the search
subspace and projected Hessian are recomputed, measured in
actor updates. Between refreshes, the stored local geometry
is reused. We vary these parameters while keeping all other
SAC and \method\ settings unchanged. Each configuration is
denoted by $r_{\max}/h/K$ and evaluated using 10 training
seeds over 300k environment interactions.
Table~\ref{tab:ant-swimmer-geometry} reports the mean AUC
and evaluation return at 300k, while
Figure~\ref{fig:exp-geometry-sensitivity} presents the
corresponding learning curves.

\begin{table}[t]
\caption{Sensitivity of \sacmethod\ to search subspace and refresh
settings on Ant and Swimmer. Entries are means $\pm$
95\% Student-$t$ confidence intervals over 10 seeds.
The configuration is given as $r_{\max}/h/K$.}
\label{tab:ant-swimmer-geometry}
\centering\small
\setlength{\tabcolsep}{3pt}
\begin{tabular}{@{}llrr@{}}
\toprule
Task & $r_{\max}/h/K$ & AUC & Return at 300k\\
\midrule
Ant & $8/16/20$ & $1016.8\pm70.4$ & $1883.0\pm254.6$\\
    & $16/32/10$ & $852.5\pm20.2$ & $1480.3\pm115.8$\\
    & $16/32/40$ & $900.9\pm53.0$ & $1615.8\pm217.6$\\
    & $32/32/50$ & $1003.3\pm60.1$ & $2232.6\pm293.2$\\
    & $32/32/64$ & $923.5\pm53.9$ & $1391.5\pm199.2$\\
\midrule
Swimmer & $32/\{32,64\}/32$ & $38.615\pm0.454$ & $40.839\pm0.351$\\
        & $32/\{32,64\}/50$ & $38.591\pm0.609$ & $40.380\pm0.499$\\
        & $32/\{32,64\}/64$ & $39.094\pm0.634$ & $41.252\pm0.395$\\
\bottomrule
\end{tabular}
\end{table}

\paragraph{Maximum subspace rank $r_{\max}$.}
Increasing the maximum subspace rank $r_{\max}$ allows more
linearly independent gradient directions to be included in
the search subspace, but configurations with higher
$r_{\max}$ do not necessarily achieve better learning performance.
On Ant, the $8/16/20$
configuration achieves the highest mean AUC,
slightly exceeding the $32/32/50$ configuration.
However, their paired AUC difference is only $13.5$,
with a 95\% confidence interval of $[-102.0,129.0]$.
Meanwhile, the latter configuration achieves a higher mean return
at 300k ($2232.6$ versus $1883.0$).
These results suggest that a relatively small search subspace
may be sufficient for achieving competitive sample efficiency.

\paragraph{History capacity $h$.}
Increasing $h$ beyond $r_{\max}-1$ does not introduce
additional historical directions into the search subspace,
regardless of the benchmark task. Consistent with this design,
$h=32$ and $h=64$ produce identical evaluation results on
Swimmer for all tested refresh intervals with $r_{\max}=32$.
Their learning curves therefore overlap in
Figure~\ref{fig:exp-geometry-sensitivity}.
This indicates that storing more historical gradients than
needed for the maximum subspace rank provides no additional
performance benefit under the current subspace construction
procedure.

\paragraph{Geometry refresh interval $K$.}
The effect of refresh frequency varies slightly across tasks and
search subspace configurations. On Ant, with $r_{\max}=32$
and $h=32$, $K=50$ achieves a higher mean AUC than
$K=64$ by $79.8$, with a paired 95\% confidence interval
of $[12.5,147.1]$. With $r_{\max}=16$ and $h=32$,
increasing $K$ from $10$ to $40$ also improves mean AUC,
although the paired confidence interval
$[-10.2,106.8]$ includes zero.

On Swimmer, with $r_{\max}=32$, $K=64$ achieves the highest
mean AUC ($39.094$), exceeding $K=32$ and $K=50$
by $0.479$ and $0.504$, respectively. Their paired
95\% confidence intervals are $[0.047,0.911]$ and
$[0.001,1.006]$. Nevertheless, the learning curves
remain close throughout much of training.

Refresh frequency also affects computational cost.
Each geometry refresh requires up to $r_{\max}$ HVPs,
giving a nominal average of $r_{\max}/K$ HVPs per actor
update. For example, with $r_{\max}=32$, increasing
$K$ from $32$ to $64$ reduces this nominal average
from one HVP to $0.5$. More frequent refreshes do
not consistently improve sample efficiency, despite
requiring additional curvature computations.

\begin{wrapfigure}{r}{0.67\textwidth}
    \centering
    \includegraphics[width=0.67\textwidth]
    {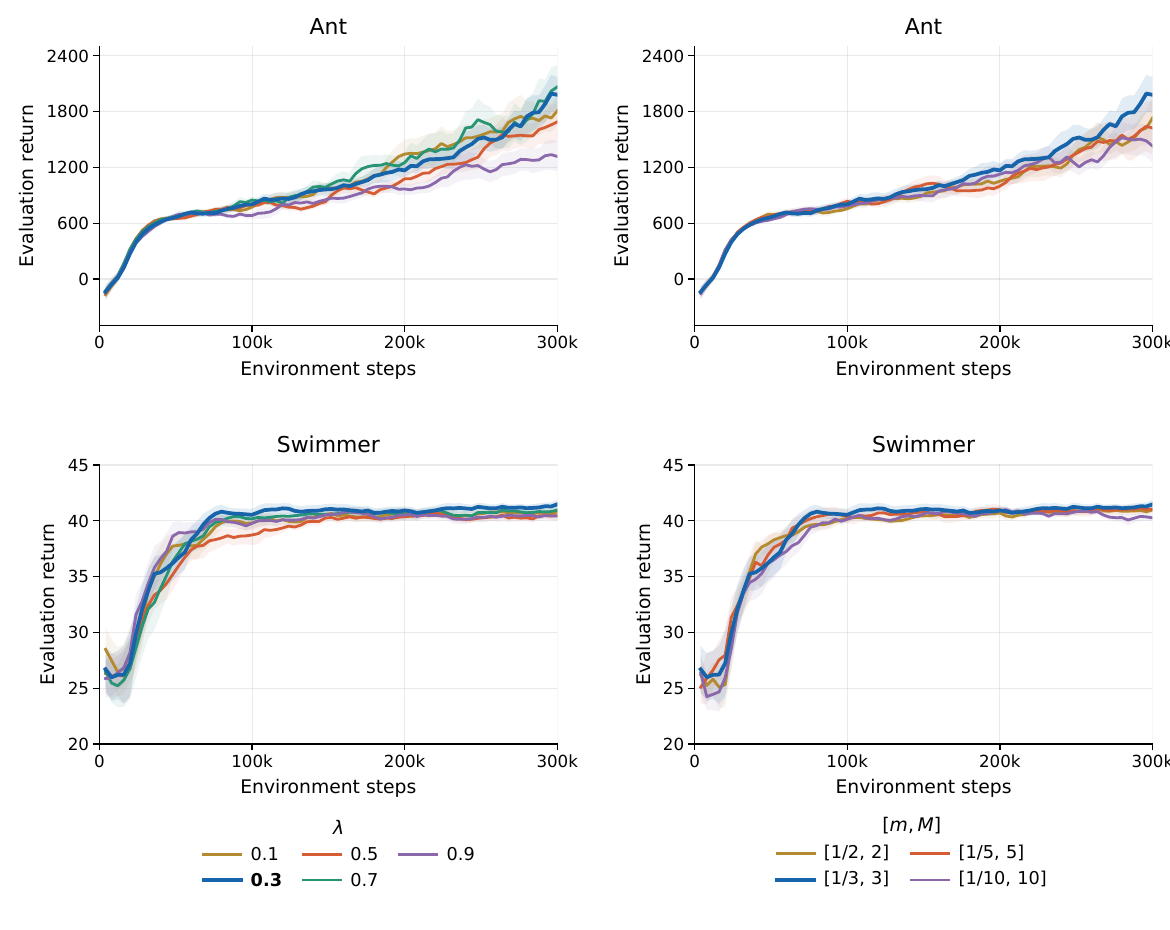}
    \caption{Sensitivity to the gradient mixing weight
    $\lambda$ (left) and spectral bounds
    $[m,M]$ (right) on Ant and Swimmer.
    Curves show mean returns, with shaded 95\% Student-$t$
    confidence intervals across 10 seeds.
    Blue curves indicate the default settings.}
    \label{fig:exp-mixing-bounds-sensitivity}
\end{wrapfigure}
\paragraph{Gradient mixing weight $\lambda$.}
The mixing weight $\lambda$ controls the contribution
of the curvature-adjusted gradient relative to the
raw gradient, ranging from no adjustment ($\lambda=0$)
to full adjustment ($\lambda=1$).
To examine its effect, we vary
$\lambda\in\{0.1,0.3,0.5,0.7,0.9\}$ on Ant and Swimmer,
using the same geometry configurations as our main
experiments and fixing the spectral bounds at $[1/3,3]$.
Each configuration is evaluated over 300k environment
interactions using 10 training seeds.

As shown in Figure~\ref{fig:exp-mixing-bounds-sensitivity},
on Ant, mean AUC varies non-monotonically with $\lambda$.
Compared with the default setting ($\lambda=0.3$),
$\lambda=0.7$ increases mean AUC by $5.2\%$,
whereas $\lambda=0.9$ reduces mean AUC by $15.6\%$
and mean return at 300k by $39.0\%$.
On Swimmer, performance differences are smaller, with
mean AUC varying by at most $2.5\%$ relative to the
default, which achieves the highest mean AUC and
return at 300k.


\paragraph{Spectral bounds $[m,M]$.}
The spectral bounds constrain the metric weights defined
in Equations~\eqref{eq:metric} and
\eqref{eq:practical-metric}, preventing excessive
amplification or suppression of gradient components
along different curvature directions.
We examine four spectral ranges, $[1/2,2]$, $[1/3,3]$,
$[1/5,5]$, and $[1/10,10]$, while fixing $\lambda=0.3$
and retaining the geometry configurations used in the
main experiments.

The default range $[1/3,3]$ achieves the highest mean
AUC among the four tested ranges on both tasks.
On Ant, alternative ranges reduce mean AUC by
approximately $6.9\%$--$8.0\%$, whereas the corresponding
reductions on Swimmer are smaller, ranging from
$0.3\%$ to $1.5\%$. Thus, neither tighter nor wider
spectral bounds consistently improve learning performance.

Further examination of the metric weights
shows that clipping occurs
primarily at the lower bound. For upper bounds of
$3$, $5$, and $10$, no upper clipping occurs, while
an upper bound of $2$ is only occasionally activated.
Therefore, the observed performance differences
mainly reflect changes in the lower bound.

Overall, the tested configurations support the use of
$r_{\max}=32$, $h=32$, refresh intervals around
$50$--$64$, $\lambda=0.3$, and spectral bounds
$[1/3,3]$ as competitive settings for \sacmethod.

\subsection{Empirical use of curvature adjusted gradients}
\label{sec:exp-mechanism}

We investigate how frequently \sacmethod\ accepts
curvature-adjusted gradients during actor training and
how this relates to policy performance. According to Algorithm~\ref{alg:sac},
\sacmethod\ requests a gradient adjustment at every actor
update. An adjustment is accepted only when the metric
satisfies the curvature-equilibration conditions in
Equation~\eqref{eq:metric-eligibility} and the mixed gradient
passes the curvature-based selection criterion in
Equation~\eqref{eq:curvature-gate}, together with the
additional numerical safeguards described in
Appendix~\ref{app:acceptance-implementation}.
Otherwise, the raw actor gradient is used.
We define the \emph{acceptance fraction} as the number of
accepted adjustments divided by the total number of requests.

Over 300k environment interactions, the acceptance
fractions are $56.6\%$ on HalfCheetah, $82.6\%$ on Walker2d,
$94.2\%$ on Swimmer, and $100.0\%$ on Pusher.
Interestingly, higher acceptance does not necessarily lead
to greater performance improvements. Pusher accepts nearly
all adjusted gradients but achieves only a $1.9\%$ mean AUC
improvement over SAC. In contrast, Walker2d achieves a much
larger improvement of $33.5\%$ despite its lower acceptance
fraction.

These findings suggest that the effectiveness of
curvature-based gradient adjustments depends not only
on their acceptance frequency, but also on
the magnitude and direction of the adjustments.
Both are influenced by the bounded spectral weights and
gradient mixing. In particular, clipping the spectral
weights may prevent exact equilibration of positive and
negative curvature. The metric eligibility conditions in
Equation~\eqref{eq:metric-eligibility} therefore require
a sufficient reduction in curvature imbalance.
Consequently, accepting more
adjusted gradients does not necessarily imply stronger
curvature equilibration or greater improvement in
actor optimization.

\subsection{Exploratory extensions to \method}
\label{sec:exp-supporting}

We also investigated several extensions to \method\ to examine
whether alternative gradient-adjustment schedules and curvature
estimation strategies could further improve actor optimization.
These experiments are detailed in
Appendix~\ref{app:exploratory-extensions}.

First, we investigate whether the frequency and timing of
curvature-based gradient adjustments affect policy optimization.
Our experiment on HalfCheetah compares two settings.
In the first, curvature-based gradient adjustments are
requested at every actor update. In the second, such
adjustments are requested only during several short,
predefined periods of training. 
Our experiments show that continuous requests can
improve sample efficiency, but require substantially
more gradient adjustments and computational effort.

Second, we investigate whether combining different spectral
weighting strategies could further improve learning performance.
We specifically develop a hybrid approach that uses \method's
curvature-equilibration weights when both positive and negative
curvature are present, and PN's inverse-absolute-curvature
weights when the projected Hessian is positive definite.
Although this hybrid design achieves higher mean AUC on Walker2d,
it does not consistently outperform the standard \method\
across all tested benchmark tasks.

Finally, we examine whether reusing historical curvature
information could improve the estimation of the projected
Hessian. Specifically, we investigate whether averaging projected Hessians
computed at different stages of actor training can improve curvature
estimation. Since the search subspace evolves during training,
historical curvature estimates can only be reused when their
subspaces are sufficiently compatible with the current one.
Our experiments show that this compatibility condition is rarely
satisfied. Hence, temporal averaging is seldom applicable and offers
little opportunity to improve learning performance.

Overall, these experiments suggest that performing
curvature-based gradient adjustments throughout training
can improve sample efficiency. In contrast, hybrid spectral
weighting and temporal averaging of projected Hessians
may not consistently provide additional performance gains.
We therefore retain the simpler \method\ design in our
main experiments.

\subsection{Summary}
\label{sec:discussion}

Our empirical findings in this section indicate that curvature
equilibration can improve the sample efficiency of SAC actor training.
Comparisons with PN as well as controlled component ablations
provide further evidence for the proposed design of \method.
Moreover, increasing the search subspace rank,
refreshing the geometry information more frequently, or averaging more
curvature minibatches does not consistently improve learning
performance. These results support a relatively simple
and computationally efficient approach to curvature estimation
for actor optimization.

In addition to all the reported experimental results,
it is worthwhile to mention that the baseline comparisons
have some limitations. In particular, PN is an adaptation
of saddle-free Newton for reinforcement learning and may
not fully reflect the potential of the original method,
which was evaluated on various other machine learning problems~\citep{dauphin2014identifying}.
Furthermore, VR-SCP uses a different policy optimization
framework and was originally tuned for MuJoCo-v2 benchmarks~\citep{khorasani2025vrscp}.
Although these benchmarks are similar to the MuJoCo-v5
tasks used in our experiments, we did not fine-tune
VR-SCP's hyper-parameters specifically for the v5 tasks.
These limitations should be considered upon interpreting
the experimental results.

\section{Conclusion}
\label{sec:conclusion}

In this paper, we introduced \method, a novel curvature-based
approach to actor optimization in deep reinforcement learning.
Unlike conventional second-order methods that primarily
rescale gradients according to curvature magnitude,
\method\ explicitly seeks to equilibrate positive and
negative curvature within a trajectory-informed search
subspace. Its key innovations include a bounded local metric
for curvature equilibration, first-order loss matching to
preserve the predicted decrease in actor loss, and curvature-based
gradient selection to determine when gradient adjustments should
be applied. Together, these components provide a principled
and practical way to exploit local curvature information
without modifying the underlying policy optimization objective.

We integrated \method\ into SAC and demonstrated its
effectiveness on 6 MuJoCo-v5 benchmarks. \sacmethod\ 
consistently achieved higher mean AUC than
SAC, PN, and VR-SCP across all 6 benchmark tasks
under the same environment interaction budget.
Further ablation and parameter sensitivity
analyses provided additional evidence for the proposed
design and demonstrated that more extensive curvature
computation does not necessarily improve learning performance.

The significance of this work extends beyond the
performance improvements achieved by \sacmethod.
Our findings demonstrate that the complex geometry of
the actor loss landscape provides valuable information
for improving policy learning.
A promising direction for future research is to make \method\ more adaptive
as actor training progresses. This could be realized by
dynamically selecting the search subspace, scheduling
curvature updates, and controlling the strength and
frequency of gradient adjustments.
Further research could also explore more efficient
curvature estimation and reuse strategies, as well as
the integration of curvature equilibration techniques into other
policy optimization algorithms.
Together, these directions offer exciting opportunities
for advancing the effectiveness, efficiency, and
applicability of geometry-aware reinforcement learning.

\clearpage
\appendix
\section{Detailed Proofs}
\label{app:proofs}
\subsection{Soft policy improvement}
\label{app:soft-proof}

\begin{proof}[Proof of Proposition~\ref{prop:soft-improvement}]
For a fixed policy $\pi$, define the soft Bellman operator
\[
 (\mathcal B_\alpha^\pi V)(s)=
 \E_{a\sim\pi(\cdot\mid s)}
 \left[R(s,a)-\alpha\log\pi(a\mid s)
 +\gamma\E_{s'\sim P(\cdot\mid s,a)}V(s')\right].
\]
This operator is monotone and a $\gamma$-contraction.
Let $q=Q_\alpha^\pi$. Under the exact policy improvement
condition, we have
\[
 \mathcal B_\alpha^{\pi'}V_\alpha^\pi
 \geq V_\alpha^\pi.
\]
Repeated application of $\mathcal B_\alpha^{\pi'}$
therefore produces a nondecreasing sequence converging
to $V_\alpha^{\pi'}$, proving
$V_\alpha^{\pi'}\geq V_\alpha^\pi$.

For approximate policy improvement, define
\[
 \mathcal I_\pi(q)(s)=
 \E_{a\sim\pi(\cdot\mid s)}
 [q(s,a)-\alpha\log\pi(a\mid s)].
\]
The critic error bound
$\norm{\widehat q-q}_\infty\leq\epsilon_Q$
implies that, for both the current policy $\pi$ and
the updated policy $\pi'$,
\[
 |\mathcal I_{\widetilde\pi}(\widehat q)(s)
 -\mathcal I_{\widetilde\pi}(q)(s)|
 \leq\epsilon_Q,\qquad
 \widetilde\pi\in\{\pi,\pi'\}.
\]

Together with the approximate improvement condition,
this gives
\[
 \begin{aligned}
 \mathcal B_\alpha^{\pi'}V_\alpha^\pi
 &=\mathcal I_{\pi'}(q)\\
 &\geq\mathcal I_{\pi'}(\widehat q)-\epsilon_Q\\
 &\geq\mathcal I_\pi(\widehat q)
       -\epsilon_Q-\epsilon_{\rm opt}\\
 &\geq V_\alpha^\pi-2\epsilon_Q-\epsilon_{\rm opt}.
 \end{aligned}
\]
Let $\epsilon=2\epsilon_Q+\epsilon_{\rm opt}$ and
$W=V_\alpha^\pi-\epsilon/(1-\gamma)$.
Since the Bellman operator satisfies
\[
 \mathcal B_\alpha^{\pi'}(V-c)
 =\mathcal B_\alpha^{\pi'}V-\gamma c
\]
for any constant $c$, we obtain
$\mathcal B_\alpha^{\pi'}W\geq W$.
By monotonicity and contraction,
$V_\alpha^{\pi'}\geq W$, proving
Equation~\eqref{eq:approx-soft-improvement}.
\end{proof}

\subsection{Local metric identities}

\begin{proof}[Proof of Proposition~\ref{prop:geometry}]
For the fixed local metric $G=B^{-1}$, the Riemannian
divergence of a vector field $X$ in the search subspace is
\[
 \operatorname{div}_G X
 =|G|^{-1/2}\sum_i
 \partial_{z_i}(|G|^{1/2}X_i)
 =\sum_i\partial_{z_i}X_i,
\]
where the second equality follows because $G$ is constant.
Setting $X=B\nabla f$, its $i$-th component is
$X_i(z)=\sum_{j=1}^r B_{ij}\partial_{z_j}f(z)$.
Since $B$ is constant, taking the divergence gives
\[
 \begin{aligned}
 \operatorname{div}_G(B\nabla f)(z)
 &=\sum_{i=1}^r\partial_{z_i}
   \left(\sum_{j=1}^r B_{ij}\partial_{z_j}f(z)\right)\\
 &=\sum_{i,j=1}^r B_{ij}
   \frac{\partial^2 f(z)}{\partial z_i\partial z_j}\\
 &=\tr(B\nabla^2f(z)),
 \end{aligned}
\]
where the last equality follows from the symmetry of the Hessian.
Evaluating at $z=0$ and using $A=\nabla^2f(0)$ yields
\[
 \operatorname{div}_G(B\nabla f)(0)=\tr(BA),
\]
proving Equation~\eqref{eq:divergence}.

Next, let $v$ be uniformly distributed on the unit sphere
$\mathbb S^{r-1}$. Since $\E[v]=0$ and
$\E[vv^\top]=I_r/r$, a Taylor expansion around $z=0$ gives
\[
 \begin{aligned}
 f(\rho B^{1/2}v)
 &=f(0)+\rho\nabla f(0)^\top B^{1/2}v\\
 &\quad+\frac{\rho^2}{2}
 v^\top B^{1/2}AB^{1/2}v+O(\rho^3).
 \end{aligned}
\]
Taking expectations, the first-order term vanishes because
$\E[v]=0$. For the quadratic term, using
$v^\top Cv=\tr(Cvv^\top)$ and $\E[vv^\top]=I_r/r$,
we obtain
\[
 \begin{aligned}
 &\E_v[v^\top B^{1/2}AB^{1/2}v]\\
 &\quad=\tr\left(B^{1/2}AB^{1/2}\E[vv^\top]\right)\\
 &\quad=\frac1r\tr(B^{1/2}AB^{1/2})\\
 &\quad=\frac1r\tr(BA),
 \end{aligned}
\]
where the last equality follows from the cyclic property
of the matrix trace. Consequently,
\[
 \E_v[f(\rho B^{1/2}v)]
 =f(0)+\frac{\rho^2}{2r}\tr(BA)+O(\rho^3),
\]
proving Equation~\eqref{eq:sphere-average}.

For the gradient update
$T_B(z)=z-\eta B\nabla f(z)$,
the Jacobian at $z=0$ is
$DT_B(0)=I_r-\eta BA$.
Since $BA$ is similar to the symmetric matrix
$B^{1/2}AB^{1/2}$, its eigenvalues are real.
For sufficiently small $\eta$, expanding the
log-determinant gives
\[
 \begin{aligned}
 \log|\det DT_B(0)|
 &=\log\det(I_r-\eta BA)\\
 &=-\eta\tr(BA)
 -\frac{\eta^2}{2}\tr((BA)^2)+O(\eta^3).
 \end{aligned}
\]
This proves Equation~\eqref{eq:volume}.
In particular, the curvature-equilibration condition
$\tr(BA)=0$ eliminates the first-order local volume
change, but higher-order changes may remain.

When the scaling factor $\chi=\chi(z)$ varies with position,
while $B$ remains fixed, applying the product rule gives
\[
 \begin{aligned}
 \operatorname{div}(\chi B\nabla f)
 &=\sum_{i=1}^r\partial_{z_i}
   \left(\chi\sum_{j=1}^r B_{ij}\partial_{z_j}f\right)\\
 &=\sum_{i,j=1}^r B_{ij}
   \left(\partial_{z_i}\chi\,\partial_{z_j}f
   +\chi\,\partial_{z_i}\partial_{z_j}f\right)\\
 &=\nabla\chi^\top B\nabla f
   +\chi\tr(B\nabla^2f),
 \end{aligned}
\]
proving Equation~\eqref{eq:scale-divergence}.

Next, consider the mixed gradient in the full actor parameter
space, with $Q$, $B$, and $\chi$ held fixed. From
Equations~\eqref{eq:matching-scale} and~\eqref{eq:direction},
the mixed gradient can be expressed as
\[
 \begin{aligned}
 u&=(1-\lambda)g+
 \lambda\left[(I_p-QQ^\top)g+\chi QBQ^\top g\right]\\
 &=\left[I_p+\lambda Q(\chi B-I_r)Q^\top\right]g\\
 &=P_\lambda g.
 \end{aligned}
\]
Since $g=\nabla\loss(\theta)$ and
$H=\nabla^2\loss(\theta)$, the divergence of this
mixed gradient field is $\tr(P_\lambda H)$ when
$P_\lambda$ is held fixed. Using the definition of
$P_\lambda$ and the cyclic property of the matrix trace,
we obtain
\[
 \begin{aligned}
 \tr(P_\lambda H)
 &=\tr(H)+\lambda\tr\left(
 Q(\chi B-I_r)Q^\top H\right)\\
 &=\tr(H)+\lambda\tr\left(
 (\chi B-I_r)Q^\top HQ\right),
 \end{aligned}
\]
proving Equation~\eqref{eq:full-trace}.

These results show that gradient rescaling and mixing can
alter the divergence of the gradient field. Consequently,
the curvature-equilibration condition $\tr(BA)=0$ does not
generally imply zero divergence for the final mixed gradient.
\end{proof}

\subsection{Feasibility and bounded metric construction}

\begin{proof}[Proof of Proposition~\ref{prop:feasibility}]
Let $A=U\diag(\mu)U^\top$ be the projected Hessian and
$B=U\diag(b)U^\top$ the inverse metric, where
$b_i\in[m,M]$. The curvature-equilibration target is
$\tr(BA)=\mu^\top b=0$.

We first determine when exact curvature equilibration is
feasible. Since $\mu^\top b=\sum_i\mu_i b_i$ is linear
in $b$, its minimum over $[m,M]^r$ is obtained by
assigning $b_i=m$ to positive eigenvalues and $b_i=M$
to negative eigenvalues. The maximum is obtained by
reversing these assignments. Thus,
\[
 \begin{aligned}
 \min_{b\in[m,M]^r}\mu^\top b&=mS_+-MS_-,\\
 \max_{b\in[m,M]^r}\mu^\top b&=MS_+-mS_-,
 \end{aligned}
\]
where $S_+=\sum_{\mu_i>0}\mu_i$ and
$S_-=\sum_{\mu_i<0}|\mu_i|$.
Because $[m,M]^r$ is convex and $\mu^\top b$ is
continuous and linear, every value between these
extrema is attainable. Therefore, exact curvature
equilibration is feasible if and only if
\[
 mS_+-MS_-\leq0\leq MS_+-mS_-,
\]
proving Equation~\eqref{eq:feasibility}.
When $S_+,S_->0$, dividing these inequalities by
the positive quantities gives
$S_+/S_-\in[m/M,M/m]$.

Next, consider the optimal bounded weights defined by
Equation~\eqref{eq:ideal-qp}. The objective
$\tfrac12\norm{b-\mathbf1}^2$ is strictly convex,
so the problem has a unique solution whenever
the equality constraint is feasible.

Introducing a Lagrange multiplier $\tau$ for
$\mu^\top b=0$, we obtain the Lagrangian
\[
 \mathcal L(b,\tau)
 =\frac12\norm{b-\mathbf1}^2+\tau\mu^\top b.
\]
For a fixed $\tau$, minimizing $\mathcal L$
over $b\in[m,M]^r$ separates into independent
problems:
\[
 b_i(\tau)=
 \argmin_{m\leq x\leq M}
 \left[\frac12(x-1)^2+\tau\mu_i x\right].
\]
The unconstrained minimum satisfies
$x-1+\tau\mu_i=0$, giving $x=1-\tau\mu_i$.
Applying the weight bounds yields
\[
 b_i(\tau)=\clip(1-\tau\mu_i,m,M).
\]

To satisfy the curvature-equilibration constraint, define
\[
 \phi(\tau)=\sum_i\mu_i
 \clip(1-\tau\mu_i,m,M).
\]
The function $\phi(\tau)$ is continuous and
nonincreasing, since each unclipped weight changes
with $\tau$ at a rate $-\mu_i$.
Furthermore,
\[
 \begin{aligned}
 \lim_{\tau\to-\infty}\phi(\tau)&=MS_+-mS_-,\\
 \lim_{\tau\to+\infty}\phi(\tau)&=mS_+-MS_-.
 \end{aligned}
\]
Under the feasibility condition, these limits
lie on opposite sides of zero or include zero.
Hence, a finite $\tau$ exists such that
$\phi(\tau)=0$. The resulting weights minimize
the Lagrangian and satisfy the equality constraint,
so they solve Equation~\eqref{eq:ideal-qp}. Their
uniqueness follows from strict convexity.
If $\mu=0$, the unique solution is $b=\mathbf1$.

Finally, consider the practical bounded weights in
Equation~\eqref{eq:practical-metric}. Let
$S=\sum_i\mu_i$, $D=\sum_i\mu_i^2$, and
$\kappa=S/(D+\epsm)$. Before clipping, the weights
are $\widetilde b_i=1-\kappa\mu_i$, giving
\[
 \begin{aligned}
 \sum_i\widetilde b_i\mu_i
 &=\sum_i\mu_i-\kappa\sum_i\mu_i^2\\
 &=S-\frac{SD}{D+\epsm}\\
 &=\frac{S\epsm}{D+\epsm}.
 \end{aligned}
\]
The implemented weights are
$b_i=\clip(\widetilde b_i,m,M)$.
Their remaining weighted curvature sum is therefore
\[
 \begin{aligned}
 \sum_i b_i\mu_i
 &=\sum_i\widetilde b_i\mu_i
   +\sum_i\mu_i(b_i-\widetilde b_i)\\
 &=\frac{S\epsm}{D+\epsm}
   +\sum_i\mu_i
   \bigl[b_i-(1-\kappa\mu_i)\bigr],
 \end{aligned}
\]
proving Equation~\eqref{eq:trace-residual}.
The first term represents the effect of denominator
regularization, while the second captures the
effect of clipping on curvature equilibration.

If $A\succeq0$ and $A\neq0$, then
\[
 \tr(BA)=\tr(B^{1/2}AB^{1/2})>0
\]
for every $B\succ0$. Similarly, if
$A\preceq0$ and $A\neq0$, then $\tr(BA)<0$.
Thus, for a nonzero projected Hessian, exact curvature
equilibration with positive weights requires both positive
and negative curvature.
\end{proof}

\subsection{First-order loss matching and gradient norm bounds}

\begin{proof}[Proof of Theorem~\ref{thm:matching}]
Let $Q^\top Q=I_r$ and $mI_r\preceq B\preceq MI_r$.
The raw gradient can be decomposed as
$g=Qa+g^\perp$, where $a=Q^\top g$ and
$g^\perp=(I_p-QQ^\top)g$. Since $Q^\top g^\perp=0$,
we have $\norm{g}^2=\norm{a}^2+\norm{g^\perp}^2$.

We first establish the first-order loss matching property.
From the definition of the adjusted gradient
$d=g^\perp+Q\chi Ba$, where
$\chi=\norm{a}^2/(a^\top Ba+\epss)$, we obtain
\[
 \begin{aligned}
 g^\top d
 &=\norm{g^\perp}^2+\chi a^\top Ba\\
 &=\norm{g^\perp}^2+
   \frac{\norm{a}^2a^\top Ba}{a^\top Ba+\epss}\\
 &=\norm{g}^2-
   \frac{\epss\norm{a}^2}{a^\top Ba+\epss}\\
 &=\norm{g}^2-\delta.
 \end{aligned}
\]
For the mixed gradient $u=(1-\lambda)g+\lambda d$,
it follows that
\[
 \begin{aligned}
 g^\top u
 &=(1-\lambda)\norm{g}^2+\lambda g^\top d\\
 &=\norm{g}^2-\lambda\delta.
 \end{aligned}
\]
Since $0\leq\delta\leq\norm{a}^2\leq\norm{g}^2$,
we obtain
$g^\top u\geq(1-\lambda)\norm{g}^2$.
Furthermore, writing $u=g+\lambda(d-g)$ gives
\[
 \begin{aligned}
 \norm{u}^2
 &=\norm{g}^2+2\lambda g^\top(d-g)
   +\lambda^2\norm{d-g}^2\\
 &=\norm{g}^2-2\lambda\delta
   +\lambda^2\norm{d-g}^2.
 \end{aligned}
\]
These identities prove the first-order matching in actor loss and
the gradient norm conditions in
Equations~\eqref{eq:matching-theorem}
and~\eqref{eq:norm-theorem}.

We next establish the upper bound on $\norm{u}$.
First, consider exact first-order matching with $\epss=0$.
Let $B=U\diag(b)U^\top$, where $b_i\in[m,M]$,
and define $\widehat a=U^\top a$. Since
$\chi=\norm{a}^2/(a^\top Ba)$, we have
\[
 \frac{\norm{\chi Ba}^2}{\norm{a}^2}
 =\frac{\norm{a}^2a^\top B^2a}{(a^\top Ba)^2}
 =\frac{\sum_i w_i b_i^2}
        {(\sum_i w_i b_i)^2},
\]
where $w_i=\widehat a_i^2/\norm{a}^2$ satisfies
$w_i\geq0$ and $\sum_iw_i=1$.

Let $x=\sum_iw_i b_i\in[m,M]$.
Since $m\leq b_i\leq M$, we have
\[
 (b_i-m)(b_i-M)\leq0,
\]
which implies
$b_i^2\leq(M+m)b_i-mM$.
Multiplying by $w_i$ and summing over $i$ gives
\[
 \sum_iw_i b_i^2\leq(M+m)x-mM.
\]
Consequently,
\[
 \begin{aligned}
 \frac{\norm{\chi Ba}^2}{\norm{a}^2}
 &\leq\frac{(M+m)x-mM}{x^2}\\
 &\leq\frac{(M+m)^2}{4mM}
 =K_B^2,
 \end{aligned}
\]
where the second inequality follows by maximizing
the expression over $x\in[m,M]$, with its maximum
attained at $x=2mM/(M+m)$.

Define the adjustment within the search subspace as
$e=\chi Ba-a$. Under exact first-order matching,
\[
 a^\top e=\chi a^\top Ba-\norm{a}^2=0.
\]
Thus, $e$ is orthogonal to $a$, and
\[
 \begin{aligned}
 \norm{e}^2
 &=\norm{\chi Ba}^2-\norm{a}^2\\
 &\leq(K_B^2-1)\norm{a}^2.
 \end{aligned}
\]
Since $d=g+Qe$, the mixed gradient becomes
$u=g+\lambda Qe$. Using
$g^\top Qe=a^\top e=0$ and $Q^\top Q=I_r$,
we obtain
\[
 \begin{aligned}
 \norm{u}^2
 &=\norm{g}^2+\lambda^2\norm{e}^2\\
 &\leq\norm{g}^2+
   \lambda^2(K_B^2-1)\norm{a}^2\\
 &\leq[1+\lambda^2(K_B^2-1)]\norm{g}^2\\
 &=C_\lambda^2\norm{g}^2.
 \end{aligned}
\]
Moreover, $\norm{u}^2\geq\norm{g}^2$, and
$g^\top u=\norm{g}^2$. Therefore, for $g\neq0$,
\[
 1\leq\frac{\norm{u}}{\norm{g}}\leq C_\lambda,
 \qquad
 \cos(g,u)=\frac{\norm{g}}{\norm{u}}
 \geq C_\lambda^{-1}.
\]
This establishes the gradient norm and alignment
bounds under exact first-order matching.

Finally, consider $\epss>0$. Define
\[
 \chi_0=\frac{\norm{a}^2}{a^\top Ba},
 \qquad
 q=\frac{a^\top Ba}{a^\top Ba+\epss}\in(0,1).
\]
Then $\chi=q\chi_0$. Using
$e=\chi_0Ba-a$, which satisfies $a^\top e=0$,
we have $\chi Ba=q(a+e)$. The mixed gradient
can therefore be written as
\[
 u=g^\perp+
 Q\bigl[(1-\lambda+\lambda q)a+\lambda q e\bigr].
\]
By orthogonality,
\[
 \begin{aligned}
 \norm{u}^2
 &=\norm{g^\perp}^2+
   (1-\lambda+\lambda q)^2\norm{a}^2 +\lambda^2q^2\norm{e}^2.
 \end{aligned}
\]
Since $0<q<1$ and $0\leq\lambda\leq1$,
both $1-\lambda+\lambda q\leq1$ and $q\leq1$.
Hence,
\[
 \begin{aligned}
 \norm{u}^2
 &\leq\norm{g^\perp}^2+\norm{a}^2
   +\lambda^2\norm{e}^2\\
 &\leq\norm{g}^2+
   \lambda^2(K_B^2-1)\norm{a}^2\\
 &\leq C_\lambda^2\norm{g}^2.
 \end{aligned}
\]
Thus, the same upper gradient norm bound holds
when the denominator regularizer $\epss>0$
is used, although exact first-order matching
is no longer guaranteed.

Meanwhile, whenever the algorithm selects the
raw gradient $u=g$, the stated bounds can also be satisfied.
\end{proof}

\subsection{Reused local geometry and actor loss comparisons}

\begin{proof}[Proof of Proposition~\ref{prop:cached-gap}]
Let $Q$ be the orthonormal search basis stored at the most
recent geometry refresh, and let
$H=\nabla^2\loss(\theta)$ be the Hessian of the current
actor loss. Since the gradient adjustment lies within
the search subspace, the raw gradient $g$ and selected
gradient $u$ share the same component outside the subspace:
\[
 g=g^\perp+Qa,\qquad u=g^\perp+Qc,
\]
where $a=Q^\top g$, $c=Q^\top u$, and
$g^\perp=(I_p-QQ^\top)g$.

Expanding the Hessian quadratic forms gives
\[
 \begin{aligned}
 u^\top Hu
 &= (g^\perp)^\top Hg^\perp
 +2(g^\perp)^\top HQc+c^\top Q^\top HQc,\\
 g^\top Hg
 &= (g^\perp)^\top Hg^\perp
 +2(g^\perp)^\top HQa+a^\top Q^\top HQa.
 \end{aligned}
\]
Subtracting these expressions, the common quadratic
term involving $g^\perp$ cancels, yielding
\[
 \begin{aligned}
 \Delta_H
 &=u^\top Hu-g^\top Hg\\
 &=c^\top Q^\top HQc-a^\top Q^\top HQa
 +2(g^\perp)^\top HQ(c-a).
 \end{aligned}
\]

Let $\overline A$ be the stored projected Hessian and
$E=Q^\top HQ-\overline A$.
Substituting $Q^\top HQ=\overline A+E$ gives
\[
 \begin{aligned}
 \Delta_H
 &=c^\top\overline A c-a^\top\overline A a +c^\top Ec-a^\top Ea
 +2(g^\perp)^\top HQ(c-a)\\
 &=\Delta_{\rm cache}+(c-a)^\top E(c+a)
 +2(g^\perp)^\top HQ(c-a),
 \end{aligned}
\]
where the last equality follows from the symmetry of
$E$. This proves Equation~\eqref{eq:cache-identity}.

Applying the operator-norm and Cauchy--Schwarz inequalities,
we have
\[
 \begin{aligned}
 |(c-a)^\top E(c+a)|
 &\leq\norm{E}\norm{c-a}\norm{c+a},\\
 |2(g^\perp)^\top HQ(c-a)|
 &\leq2\norm{Q^\top Hg^\perp}\norm{c-a}.
 \end{aligned}
\]
Combining these inequalities with
Equation~\eqref{eq:cache-identity} yields
\[
 \begin{aligned}
 \Delta_H\leq\Delta_{\rm cache}
 &+\norm{E}\norm{c-a}\norm{c+a} +2\norm{Q^\top Hg^\perp}\norm{c-a},
 \end{aligned}
\]
proving Equation~\eqref{eq:cache-bound}.

At a geometry refresh, if the stored projected Hessian
is computed from the current actor loss without numerical
error, then $\overline A=Q^\top HQ$ and hence $E=0$.
Since the refreshed search basis contains the current raw
gradient $g$, we also have $g^\perp=0$.
Consequently, $\Delta_H=\Delta_{\rm cache}$, so accepting
a mixed gradient with $\Delta_{\rm cache}<0$ guarantees
$\Delta_H<0$.

Between refreshes, a sufficient condition for
$\Delta_H<0$ is
\[
 -\Delta_{\rm cache}>
 \norm{E}\norm{c-a}\norm{c+a}
 +2\norm{Q^\top Hg^\perp}\norm{c-a}.
\]
Thus, the reduction in the Hessian quadratic form
measured using the stored projected Hessian must
outweigh the effects of changes in projected curvature
and interactions with the gradient component outside
the search subspace.
\end{proof}

\begin{proof}[Proof of Corollary~\ref{cor:local-comparison}]
Let $g=\nabla\loss(\theta)$,
$H=\nabla^2\loss(\theta)$, and let $u$ be the
accepted mixed gradient. Assume that the Hessian of
$\loss$ is $L_H$-Lipschitz continuous along the
line segments from $\theta$ to $\theta-\eta u$
and to $\theta-\eta g$.

For either update direction $v\in\{u,g\}$,
a second-order Taylor expansion around $\theta$ gives
\[
 \loss(\theta-\eta v)
 =\loss(\theta)-\eta g^\top v
 +\frac{\eta^2}{2}v^\top Hv+R_v,
\]
where the Taylor remainder satisfies
\[
 |R_v|\leq\frac{L_H\eta^3}{6}\norm{v}^3.
\]

Applying this expansion to the mixed-gradient and
raw-gradient updates, then subtracting, yields
\[
 \begin{aligned}
 &\loss(\theta-\eta u)-\loss(\theta-\eta g)\\
 &=-\eta g^\top(u-g)
 +\frac{\eta^2}{2}(u^\top Hu-g^\top Hg)
 +R_u-R_g.
 \end{aligned}
\]

From Theorem~\ref{thm:matching}, first-order loss
matching gives
\[
 g^\top u=\norm{g}^2-\lambda\delta.
\]
Therefore,
\[
 -\eta g^\top(u-g)=\eta\lambda\delta.
\]
By the definition of $\Delta_H$,
\[
 u^\top Hu-g^\top Hg=\Delta_H.
\]
Finally, using the Taylor remainder bounds,
\[
 \begin{aligned}
 R_u-R_g
 &\leq |R_u|+|R_g|\\
 &\leq\frac{L_H\eta^3}{6}
 \bigl(\norm{u}^3+\norm{g}^3\bigr).
 \end{aligned}
\]
Combining these results gives
\[
 \begin{aligned}
 \loss(\theta-\eta u)-\loss(\theta-\eta g)
 &\leq\eta\lambda\delta+\frac{\eta^2}{2}\Delta_H +\frac{L_H\eta^3}{6}
 \bigl(\norm{u}^3+\norm{g}^3\bigr),
 \end{aligned}
\]
proving Equation~\eqref{eq:local-comparison}.

For a quadratic actor loss, the Taylor remainders
vanish. Under exact first-order matching
($\epss=0$), we also have $\delta=0$.
The loss difference therefore reduces to
\[
 \loss(\theta-\eta u)-\loss(\theta-\eta g)
 =\frac{\eta^2}{2}\Delta_H.
\]
Hence, $\Delta_H<0$ guarantees that the mixed-gradient
update achieves a lower actor loss than the raw-gradient
update for any $\eta>0$.

For a nonquadratic actor loss, the same conclusion
holds whenever
\[
 -\frac{\eta^2}{2}\Delta_H
 >
 \eta\lambda\delta+
 \frac{L_H\eta^3}{6}
 \bigl(\norm{u}^3+\norm{g}^3\bigr).
\]
This condition shows that the reduction in the
Hessian quadratic form must outweigh the Taylor
remainder and any first-order matching error.
\end{proof}

\subsection{Convergence for a fixed actor loss and accumulated improvement}

\begin{proof}[Proof of Theorem~\ref{thm:stationary}]
Let $g_t=\nabla\loss(\theta_t)$ and consider the direct
actor update $\theta_{t+1}=\theta_t-\eta u_t$.
Since $\loss$ has an $L$-Lipschitz gradient, the
descent lemma gives
\[
 \loss(\theta_{t+1})
 \leq\loss(\theta_t)-\eta g_t^\top u_t
 +\frac{L\eta^2}{2}\norm{u_t}^2.
\]
By Assumption~\ref{ass:direct}, the selected gradient
satisfies
$g_t^\top u_t\geq c_0\norm{g_t}^2$ and
$\norm{u_t}\leq C_0\norm{g_t}$.
Substituting these bounds yields
\[
 \begin{aligned}
 \loss(\theta_{t+1})
 &\leq\loss(\theta_t)-\eta c_0\norm{g_t}^2
 +\frac{L\eta^2C_0^2}{2}\norm{g_t}^2\\
 &=\loss(\theta_t)-a_\eta\norm{g_t}^2,
 \end{aligned}
\]
where $a_\eta=\eta c_0-L\eta^2C_0^2/2>0$
for $0<\eta<2c_0/(LC_0^2)$.

Summing over $t=0,\ldots,T-1$, we obtain
\[
 \begin{aligned}
 a_\eta\sum_{t=0}^{T-1}\norm{g_t}^2
 &\leq\sum_{t=0}^{T-1}
 \left[\loss(\theta_t)-\loss(\theta_{t+1})\right]\\
 &=\loss(\theta_0)-\loss(\theta_T)\\
 &\leq\loss(\theta_0)-\loss_*,
 \end{aligned}
\]
where the last inequality follows from the lower bound
$\loss(\theta_T)\geq\loss_*$.
Dividing by $Ta_\eta$ and using the fact that
the minimum is no greater than the average gives
\[
 \min_{0\leq t<T}\norm{g_t}^2
 \leq\frac1T\sum_{t=0}^{T-1}\norm{g_t}^2
 \leq\frac{\loss(\theta_0)-\loss_*}{Ta_\eta},
\]
proving Equation~\eqref{eq:stationary-rate}.

Next, suppose $\loss$ satisfies the
Polyak--\L{}ojasiewicz (PL) inequality
\[
 \frac12\norm{g_t}^2
 \geq\mu_{\rm PL}\bigl(\loss(\theta_t)-\loss_*\bigr).
\]
Combining this inequality with the descent bound gives
\[
 \begin{aligned}
 \loss(\theta_{t+1})-\loss_*
 &\leq\loss(\theta_t)-\loss_*-a_\eta\norm{g_t}^2\\
 &\leq(1-2\mu_{\rm PL}a_\eta)
 \bigl(\loss(\theta_t)-\loss_*\bigr).
 \end{aligned}
\]
When $0\leq1-2\mu_{\rm PL}a_\eta<1$,
repeated application yields
\[
 \loss(\theta_T)-\loss_*
 \leq(1-2\mu_{\rm PL}a_\eta)^T
 \bigl(\loss(\theta_0)-\loss_*\bigr),
\]
proving the conditional linear convergence rate in
Equation~\eqref{eq:pl-rate}.
\end{proof}

\begin{proof}[Proof of Proposition~\ref{prop:surplus}]
Let $g_t=\nabla\loss(\theta_t)$.
For a direct update using the raw gradient,
the descent lemma gives
\[
 \begin{aligned}
 \loss(\theta_t-\eta g_t)
 &\leq\loss(\theta_t)-\eta\norm{g_t}^2
 +\frac{L\eta^2}{2}\norm{g_t}^2\\
 &=\loss(\theta_t)
 -\left(\eta-\frac{L\eta^2}{2}\right)\norm{g_t}^2.
 \end{aligned}
\]
Since $0<\eta<2/L$, the coefficient
$\eta-L\eta^2/2$ is positive.

By the additional loss improvement condition in
Proposition~\ref{prop:surplus}, the selected actor update satisfies
\[
 \loss(\theta_{t+1})
 \leq\loss(\theta_t-\eta g_t)-I_ts_t,
\]
where $I_t$ indicates whether the mixed gradient is
accepted and $s_t\geq0$ denotes its additional loss decrease
relative to the raw-gradient update.

Combining the two inequalities gives
\[
 \loss(\theta_{t+1})
 \leq\loss(\theta_t)
 -\left(\eta-\frac{L\eta^2}{2}\right)\norm{g_t}^2
 -I_ts_t.
\]
Rearranging and summing over $t=0,\ldots,T-1$,
we obtain
\[
 \begin{aligned}
 &\left(\eta-\frac{L\eta^2}{2}\right)
 \sum_{t=0}^{T-1}\norm{g_t}^2
 +\sum_{t=0}^{T-1}I_ts_t\\
 &\quad\leq
 \sum_{t=0}^{T-1}
 \left[\loss(\theta_t)-\loss(\theta_{t+1})\right]\\
 &\quad=\loss(\theta_0)-\loss(\theta_T)\\
 &\quad\leq\loss(\theta_0)-\loss_*.
 \end{aligned}
\]
This proves Equation~\eqref{eq:surplus}.
The quantity $\sum_{t=0}^{T-1}I_ts_t$
measures the accumulated additional decrease in actor loss
obtained by using accepted mixed gradients along the actor
optimization trajectory.
\end{proof}

\subsection{Convergence for changing actor losses}

\begin{proof}[Proof of Theorem~\ref{thm:moving}]
Let $\bar{\loss}_t$ denote the conditional expected actor
loss and $\bar g_t=\nabla\bar{\loss}_t(\theta_t)$ its gradient.
Consider the direct actor update
$\theta_{t+1}=\theta_t-\eta u_t$.

Since $\bar{\loss}_t$ has an $L$-Lipschitz gradient,
the descent lemma gives
\[
 \bar{\loss}_t(\theta_{t+1})
 \leq\bar{\loss}_t(\theta_t)
 -\eta\bar g_t^\top u_t
 +\frac{L\eta^2}{2}\norm{u_t}^2.
\]
Taking conditional expectations with respect to
$\mathcal F_t$, we obtain
\[
 \begin{aligned}
 &\E[\bar{\loss}_t(\theta_{t+1})\mid\mathcal F_t]
 \leq\bar{\loss}_t(\theta_t)
 -\eta\E[\bar g_t^\top u_t\mid\mathcal F_t]+\frac{L\eta^2}{2}
 \E[\norm{u_t}^2\mid\mathcal F_t].
 \end{aligned}
\]
By Assumption~\ref{ass:moving}, the expected gradient
alignment and squared norm satisfy
\[
 \begin{aligned}
 \E[\bar g_t^\top u_t\mid\mathcal F_t]
 &\geq c_0\norm{\bar g_t}^2-\beta_t,\\
 \E[\norm{u_t}^2\mid\mathcal F_t]
 &\leq C_0^2\norm{\bar g_t}^2+\sigma_t^2.
 \end{aligned}
\]
Substituting these bounds gives
\[
 \begin{aligned}
 &\E[\bar{\loss}_t(\theta_{t+1})\mid\mathcal F_t]\\
 &\quad\leq\bar{\loss}_t(\theta_t)
 -\left(\eta c_0-\frac{L\eta^2C_0^2}{2}\right)
 \norm{\bar g_t}^2+\eta\beta_t+\frac{L\eta^2}{2}\sigma_t^2\\
 &\quad=\bar{\loss}_t(\theta_t)
 -a_\eta\norm{\bar g_t}^2
 +\eta\beta_t+\frac{L\eta^2}{2}\sigma_t^2,
 \end{aligned}
\]
where $a_\eta=\eta c_0-L\eta^2C_0^2/2>0$.

Next, the expected actor loss may change between
updates. By Equation~\eqref{eq:objective-drift},
\[
 \begin{aligned}
 &\E[\bar{\loss}_{t+1}(\theta_{t+1})\mid\mathcal F_t]
 \leq \E[\bar{\loss}_t(\theta_{t+1})\mid\mathcal F_t]+v_t.
 \end{aligned}
\]
Combining this inequality with the preceding descent
bound and taking full expectations yields
\[
 \begin{aligned}
 \E[\bar{\loss}_{t+1}(\theta_{t+1})]
 &\leq\E[\bar{\loss}_t(\theta_t)]
 -a_\eta\E\norm{\bar g_t}^2 +\E\left[
 v_t+\eta\beta_t+\frac{L\eta^2}{2}\sigma_t^2
 \right].
 \end{aligned}
\]

Summing over $t=0,\ldots,T-1$, the expected actor
loss values telescope:
\[
 \begin{aligned}
 a_\eta\sum_{t=0}^{T-1}\E\norm{\bar g_t}^2
 &\leq\E[\bar{\loss}_0(\theta_0)]
 -\E[\bar{\loss}_T(\theta_T)] +\sum_{t=0}^{T-1}
 \E\left[v_t+\eta\beta_t
 +\frac{L\eta^2}{2}\sigma_t^2\right].
 \end{aligned}
\]
Since $\bar{\loss}_T(\theta_T)\geq\loss_*$,
dividing by $Ta_\eta$ gives
\[
 \frac1T\sum_{t=0}^{T-1}\E\norm{\bar g_t}^2
 \leq
 \frac{
 \E[\bar{\loss}_0(\theta_0)]-\loss_*
 +\sum_{t=0}^{T-1}\E[
 v_t+\eta\beta_t+(L\eta^2/2)\sigma_t^2]
 }{Ta_\eta},
\]
proving Equation~\eqref{eq:moving-rate}.

For a fixed expected actor loss with
$\beta_t=v_t=0$ and uniformly bounded $\sigma_t^2$,
the bound simplifies to
\[
 \frac1T\sum_{t=0}^{T-1}\E\norm{\bar g_t}^2
 =O\left(\frac{1}{T\eta}+\eta\right),
\]
for sufficiently small $\eta$.
Choose $\eta$ proportional to $T^{-1/2}$, with
$a_\eta>0$. Then $1/(T\eta)$ and $\eta$ are both
$O(T^{-1/2})$, giving the stated stationarity bound.
\end{proof}

\begin{proof}[Derivation of Equation~\eqref{eq:conservative-bias}]
Let $g_t$ be the sampled raw actor gradient and
$\bar g_t$ its conditional expectation.
Under conditional unbiasedness, write
\[
 g_t=\bar g_t+\zeta_t,
 \qquad
 \E[\zeta_t\mid\mathcal F_t]=0,
\]
where $\zeta_t$ denotes the gradient sampling error.
Assume that
\[
 \E[\norm{\zeta_t}^2\mid\mathcal F_t]
 \leq\sigma_{g,t}^2.
\]

The selected gradient can be written as
$u_t=g_t+h_t$, where $h_t$ is the gradient
adjustment. Under exact first-order matching,
$g_t^\top h_t=0$. Therefore,
\[
 \begin{aligned}
 &\bar g_t^\top\E[u_t\mid\mathcal F_t]\\
 &\quad=\bar g_t^\top\E[g_t+h_t\mid\mathcal F_t]\\
 &\quad=\norm{\bar g_t}^2
 +\E[(g_t-\zeta_t)^\top h_t\mid\mathcal F_t]\\
 &\quad=\norm{\bar g_t}^2
 -\E[\zeta_t^\top h_t\mid\mathcal F_t].
 \end{aligned}
\]

By Theorem~\ref{thm:matching}, the adjustment satisfies
$\norm{h_t}\leq\kappa_h\norm{g_t}$, where
$\kappa_h=\lambda\sqrt{K_B^2-1}$.
Applying the conditional Cauchy--Schwarz inequality gives
\[
 \begin{aligned}
 &\left|\E[\zeta_t^\top h_t\mid\mathcal F_t]\right|\\
 &\quad\leq
 \sqrt{\E[\norm{\zeta_t}^2\mid\mathcal F_t]}
 \sqrt{\E[\norm{h_t}^2\mid\mathcal F_t]}\\
 &\quad\leq\kappa_h\sigma_{g,t}
 \sqrt{\E[\norm{g_t}^2\mid\mathcal F_t]}.
 \end{aligned}
\]
Since $g_t=\bar g_t+\zeta_t$ and
$\E[\zeta_t\mid\mathcal F_t]=0$, we have
\[
 \begin{aligned}
 \E[\norm{g_t}^2\mid\mathcal F_t]
 &=\norm{\bar g_t}^2+
 \E[\norm{\zeta_t}^2\mid\mathcal F_t]\\
 &\leq\norm{\bar g_t}^2+\sigma_{g,t}^2.
 \end{aligned}
\]
Consequently,
\[
 \begin{aligned}
 &\bar g_t^\top\E[u_t\mid\mathcal F_t]\\
 &\quad\geq\norm{\bar g_t}^2
 -\kappa_h\sigma_{g,t}
 \sqrt{\norm{\bar g_t}^2+\sigma_{g,t}^2}\\
 &\quad\geq\norm{\bar g_t}^2
 -\kappa_h\sigma_{g,t}\norm{\bar g_t}
 -\kappa_h\sigma_{g,t}^2,
 \end{aligned}
\]
where the last inequality uses
$\sqrt{x^2+y^2}\leq x+y$ for $x,y\geq0$.

For any $\epsilon\in(0,1)$, Young's inequality gives
\[
 \kappa_h\sigma_{g,t}\norm{\bar g_t}
 \leq\epsilon\norm{\bar g_t}^2
 +\frac{\kappa_h^2}{4\epsilon}\sigma_{g,t}^2.
\]
Substituting this bound yields
\[
 \begin{aligned}
 \E[\bar g_t^\top u_t\mid\mathcal F_t]
 &\geq(1-\epsilon)\norm{\bar g_t}^2-
 \left(\kappa_h+\frac{\kappa_h^2}{4\epsilon}\right)
 \sigma_{g,t}^2,
 \end{aligned}
\]
proving Equation~\eqref{eq:conservative-bias}.

Finally, Theorem~\ref{thm:matching} also gives
$\norm{u_t}\leq C_\lambda\norm{g_t}$.
Taking conditional expectations,
\[
 \begin{aligned}
 \E[\norm{u_t}^2\mid\mathcal F_t]
 &\leq C_\lambda^2
 \E[\norm{g_t}^2\mid\mathcal F_t]\\
 &\leq C_\lambda^2
 (\norm{\bar g_t}^2+\sigma_{g,t}^2).
 \end{aligned}
\]
Thus, the expected alignment and norm bounds in
Assumption~\ref{ass:moving} can be satisfied with
$c_0=1-\epsilon$,
$\beta_t=(\kappa_h+\kappa_h^2/(4\epsilon))
\sigma_{g,t}^2$,
$C_0=C_\lambda$, and
$\sigma_t^2=C_\lambda^2\sigma_{g,t}^2$.
\end{proof}

\subsection{From actor loss optimization to the soft policy gradient}

\begin{proof}[Proof of Proposition~\ref{prop:policy-bridge}]
We first establish the relationship between the exact actor
gradient and the entropy-regularized return $J_\alpha(\theta)$
defined in Equation~\eqref{eq:soft-objective}.

From Equation~\eqref{eq:soft-values}, the soft state-value
function satisfies
\[
 V_\alpha^{\pi_\theta}(s)
 =\E_{a\sim\pi_\theta(\cdot\mid s)}
 \left[Q_\alpha^{\pi_\theta}(s,a)
 -\alpha\log\pi_\theta(a\mid s)\right].
\]
Differentiating with respect to $\theta$ gives
\[
 \begin{aligned}
 \nabla_\theta V_\alpha^{\pi_\theta}(s)
 &=\E_{a\sim\pi_\theta}
 \Bigl[
 \nabla_\theta\log\pi_\theta(a\mid s)
 \bigl(Q_\alpha^{\pi_\theta}(s,a)
 -\alpha\log\pi_\theta(a\mid s)\bigr)
 \Bigr] +\E_{a\sim\pi_\theta}
 \left[\nabla_\theta Q_\alpha^{\pi_\theta}(s,a)\right],
 \end{aligned}
\]
where the additional term
$-\alpha\E_{a\sim\pi_\theta}
[\nabla_\theta\log\pi_\theta(a\mid s)]$
vanishes because the policy density integrates to one.
To simplify the derivation, define the exact state-wise
actor-loss gradient
\[
 \begin{aligned}
 \psi_\theta^*(s)
 =-\E_{a\sim\pi_\theta}
 \Bigl[
 &\nabla_\theta\log\pi_\theta(a\mid s) \cdot\bigl(Q_\alpha^{\pi_\theta}(s,a)
 -\alpha\log\pi_\theta(a\mid s)\bigr)
 \Bigr].
 \end{aligned}
\]
From the soft action-value definition,
\[
 Q_\alpha^{\pi_\theta}(s,a)
 =R(s,a)+\gamma\E_{s'\sim P(\cdot\mid s,a)}
 V_\alpha^{\pi_\theta}(s'),
\]
we have
\[
 \nabla_\theta Q_\alpha^{\pi_\theta}(s,a)
 =\gamma\E_{s'\sim P(\cdot\mid s,a)}
 \nabla_\theta V_\alpha^{\pi_\theta}(s').
\]
Consequently,
\[
 \begin{aligned}
 \nabla_\theta V_\alpha^{\pi_\theta}(s)
 &=-\psi_\theta^*(s) +\gamma\E_{a\sim\pi_\theta,\,
 s'\sim P(\cdot\mid s,a)}
 \nabla_\theta V_\alpha^{\pi_\theta}(s').
 \end{aligned}
\]
Recursively expanding this equation and taking the
expectation over the initial state distribution $\rho_0$
yields
\[
 \begin{aligned}
 \nabla J_\alpha(\theta)
 &=-\sum_{k=0}^{\infty}\gamma^k
 \E_{s_k\sim p_{\theta,k}}[\psi_\theta^*(s_k)]\\
 &=-\frac{1}{1-\gamma}
 \E_{s\sim d_\theta}[\psi_\theta^*(s)],
 \end{aligned}
\]
where $d_\theta$ is the normalized discounted state
distribution in Equation~\eqref{eq:occupancy}.

With the exact critic held fixed, the expected actor loss
can be expressed using either $a\sim\pi_\theta(\cdot\mid s)$
or $a=a_\theta(s,\xi)$. Differentiating the first
expression gives $\psi_\theta^*(s)$. For the second,
the assumed interchange of differentiation and
expectation gives
\[
 \begin{aligned}
 \psi_\theta^*(s)
 =\E_\xi\nabla_\theta\Bigl[
 &\alpha\log\pi_\theta(a_\theta(s,\xi)\mid s)
 -Q_\alpha^{\pi_\theta}(s,a_\theta(s,\xi))
 \Bigr]_{Q_\mathrm{fixed}}.
 \end{aligned}
\]
Here, we treat the exact critic as fixed while allowing gradients to propagate
through its action input to the actor. The effect of policy changes on future
values is already accounted for by the preceding Bellman recursion.
Therefore, the exact actor gradient satisfies
\[
 \begin{aligned}
 g_\theta^*
 &=\E_{s\sim d_\theta}[\psi_\theta^*(s)]\\
 &=-(1-\gamma)\nabla J_\alpha(\theta),
 \end{aligned}
\]
proving Equation~\eqref{eq:soft-pg-bridge}.

Next, we relate the expected actor gradient $\bar g_t$
to the exact actor gradient $g_{\theta_t}^*$.
By the definition of the gradient error $e_t$,
\[
 \norm{\bar g_t-g_{\theta_t}^*}=e_t.
\]
Since
$g_{\theta_t}^*=-(1-\gamma)\nabla J_\alpha(\theta_t)$,
we obtain
\[
 \begin{aligned}
 &(1-\gamma)^2
 \norm{\nabla J_\alpha(\theta_t)}^2\\
 &\quad=\norm{g_{\theta_t}^*}^2\\
 &\quad=\norm{\bar g_t-
 (\bar g_t-g_{\theta_t}^*)}^2\\
 &\quad\leq2\norm{\bar g_t}^2+2e_t^2,
 \end{aligned}
\]
where the last inequality uses
$\norm{x+y}^2\leq2\norm{x}^2+2\norm{y}^2$.
Taking expectations and averaging over $T$ actor updates
gives
\[
 \frac1T\sum_{t<T}\E [ \norm{\nabla J_\alpha(\theta_t)}^2 ]
 \leq\frac{2}{(1-\gamma)^2}
 \left[
 \frac1T\sum_{t<T}\E [ \norm{\bar g_t}^2 ]
 +\frac1T\sum_{t<T}\E [ e_t^2 ]
 \right],
\]
proving Equation~\eqref{eq:return-stationarity}.

Finally, we establish an upper bound on $e_t$.
For a fixed critic $q$, define the state-wise actor
gradient
\[
 \psi_\theta(s;q)=
 \E_\xi\nabla_\theta
 \left[
 \alpha\log\pi_\theta(a_\theta(s,\xi)\mid s)
 -q(s,a_\theta(s,\xi))
 \right]_{q_\mathrm{fixed}}.
\]
Let $\widehat Q_t$ denote the learned critic used in
the SAC actor loss. Then
\[
 \begin{aligned}
 \bar g_t
 &=\E_{s\sim\nu_t^{\rm replay}}
 [\psi_{\theta_t}(s;\widehat Q_t)],\\
 g_{\theta_t}^*
 &=\E_{s\sim d_{\theta_t}}
 [\psi_{\theta_t}(s;Q_\alpha^{\pi_{\theta_t}})].
 \end{aligned}
\]
Adding and subtracting the expectation of the exact
state-wise gradient under the replay distribution,
then applying the triangle inequality, gives
\[
 \begin{aligned}
 e_t
 &\leq
 \Bigl\|
 \E_{s\sim\nu_t^{\rm replay}}
 [\psi_{\theta_t}(s;\widehat Q_t)
 -\psi_{\theta_t}(s;Q_\alpha^{\pi_{\theta_t}})]
 \Bigr\|\\
 &\quad+
 \Bigl\|
 \E_{s\sim\nu_t^{\rm replay}}
 [\psi_{\theta_t}(s;Q_\alpha^{\pi_{\theta_t}})]
 -\E_{s\sim d_{\theta_t}}
 [\psi_{\theta_t}(s;Q_\alpha^{\pi_{\theta_t}})]
 \Bigr\|.
 \end{aligned}
\]

The first quantity measures the effect of critic
approximation. Since the entropy term is identical
for both critics, their state-wise gradient difference
depends only on the action-value gradients.
Under the assumed bounds
$\norm{D_\theta a_\theta}\leq A_\theta$ and
\[
 \norm{\nabla_a\widehat Q_t(s,a)
 -\nabla_a Q_\alpha^{\pi_{\theta_t}}(s,a)}
 \leq\epsilon_{Q,t},
\]
we have
\[
 \begin{aligned}
 &\norm{\psi_{\theta_t}(s;\widehat Q_t)
 -\psi_{\theta_t}(s;Q_\alpha^{\pi_{\theta_t}})}\\
 &\quad\leq
 \E_\xi\left[
 \norm{D_\theta a_{\theta_t}}
 \norm{\nabla_a\widehat Q_t
 -\nabla_a Q_\alpha^{\pi_{\theta_t}}}
 \right]\\
 &\quad\leq A_\theta\epsilon_{Q,t}.
 \end{aligned}
\]

The second quantity measures the effect of replay
distribution mismatch. Since
$\norm{\psi_{\theta_t}(s;Q_\alpha^{\pi_{\theta_t}})}
\leq G_\psi$ for every state, the total variation
inequality gives
\[
 \begin{aligned}
 &\Bigl\|
 \E_{s\sim\nu_t^{\rm replay}}
 [\psi_{\theta_t}(s;Q_\alpha^{\pi_{\theta_t}})]
 -\E_{s\sim d_{\theta_t}}
 [\psi_{\theta_t}(s;Q_\alpha^{\pi_{\theta_t}})]
 \Bigr\| \leq
 2G_\psi\TV(\nu_t^{\rm replay},d_{\theta_t}).
 \end{aligned}
\]
Combining the two bounds yields
\[
 e_t\leq
 2G_\psi\TV(\nu_t^{\rm replay},d_{\theta_t})
 +A_\theta\epsilon_{Q,t},
\]
proving Equation~\eqref{eq:replay-critic-bias}.
For SAC, $\widehat Q_t$ is the minimum of the two
learned critics (i.e., action-value functions). The derivation
applies where this minimum is differentiable, provided that both
critics satisfy the assumed action-gradient error bound.
\end{proof}

\section{Implementation Details}
\label{app:implementation}

This appendix describes the numerical checks and geometry-refresh rules
used to implement \method\ in SAC. \sacmethod\ keeps the original SAC
actor loss, critic updates, replay buffer, and Adam optimizer. It changes
only the gradient supplied to the actor optimizer, as defined in
Equation~\eqref{eq:accepted-input}. The search subspace is constructed
from recent gradients supplied to the actor optimizer, and the projected
Hessian is computed using the same actor loss and fixed actor noise as
the current gradient, with the critics held fixed. The implementation
uses double precision for the small projected-Hessian eigendecomposition,
while the actor and its optimizer retain their usual precision.

\subsection{Numerical first-order loss matching}
\label{app:matching-implementation}

Let $g_t$ be the current raw actor gradient, and let $Q$, $U$,
$\mu$, and $b$ denote the search basis, curvature eigenvectors,
eigenvalues, and metric weights stored at the most recent geometry
refresh. To implement Equation~\eqref{eq:matching-scale}, we compute
\begin{equation}
\begin{gathered}
 z_t=Q^\top g_t,\qquad a_t=U^\top z_t,\qquad
 v_t=b\odot a_t,\\
 \chi_t=\frac{\norm{a_t}^2}{a_t^\top v_t+\epss},
\end{gathered}
\label{eq:regularized-matching-scale}
\end{equation}
where $\odot$ denotes elementwise multiplication and
$\epss=10^{-12}$ prevents division by zero. The adjusted gradient is
\(
 d_t=g_t+Q\bigl(U(\chi_tv_t)-z_t\bigr).
\)
We require $a_t^\top v_t>10^{-12}$, a finite scaling factor
$0<\chi_t\leq10$, and
\begin{equation}
 \left|g_t^\top d_t-\norm{g_t}^2\right|
 \leq10^{-5}\bigl(1+\norm{g_t}^2\bigr).
 \label{eq:matching-check}
\end{equation}
These numerical stability checks ensure that the adjusted gradient approximately
preserves the predicted first-order decrease in actor loss. If any
check fails, we use the raw gradient $g_t$ instead to update the actor.

\subsection{Curvature-based gradient selection and safety checks}
\label{app:acceptance-implementation}

When the local metric satisfies Equation~\eqref{eq:metric-eligibility}
and first-order matching succeeds, we form the mixed gradient
$\widetilde u_t$ using Equation~\eqref{eq:direction}. Its coordinates
in the stored curvature eigenbasis are
\(
 c_t=(1-\lambda)a_t+\lambda\chi_tv_t.
\)
The two curvature terms in Equation~\eqref{eq:curvature-gate}
can therefore be calculated as
\begin{equation}
 K_t^{\rm raw}=\sum_i\mu_i a_{t,i}^2,\qquad
 K_t^{\rm mix}=\sum_i\mu_i c_{t,i}^2.
 \label{eq:curvature-eigencoordinates}
\end{equation}
We select the mixed gradient only if it satisfies the numerical
margin:
\begin{equation}
 K_t^{\rm mix}-K_t^{\rm raw}
 +10^{-4}\left(\sum_i|\mu_i|a_{t,i}^2+10^{-12}\right)<0.
 \label{eq:curvature-margin}
\end{equation}
This criterion requires $K_t^{\rm mix}$ to be lower than
$K_t^{\rm raw}$ by a numerical margin. With exact
first-order loss matching ($\epss=0$), the stored
projected quadratic model predicts a lower loss for
the direct mixed-gradient update. With regularized
matching ($\epss>0$), a lower predicted loss also
requires the curvature reduction to outweigh the
first-order matching error.
Corollary~\ref{cor:local-comparison} gives the
corresponding loss comparison for direct actor updates.

The mixed gradient must also be finite and satisfy
\begin{equation}
\begin{aligned}
 \frac{g_t^\top\widetilde u_t}
 {\norm{g_t}\norm{\widetilde u_t}+10^{-12}}&\geq0.05,\\
 0.9\leq\frac{\norm{\widetilde u_t}}
 {\norm{g_t}+10^{-12}}&\leq1.1,\\
 \frac{g_t^\top\widetilde u_t}
 {\norm{g_t}^2+10^{-12}}&\geq0.05.
\end{aligned}
\label{eq:input-gates}
\end{equation}
These checks limit changes in gradient direction and magnitude.
The final norm bound is checked again before the actor update.
If the metric, matching, curvature-selection, or safety checks fail,
Adam receives $u_t=g_t$; otherwise, it receives
$u_t=\widetilde u_t$. These criteria govern gradient selection
using the stored local geometry. They do not guarantee a lower
actor loss after an Adam update.

\subsection{Geometry refresh and gradient history}
\label{app:geometry-reuse}

\sacmethod\ requests
a gradient adjustment at every actor update. The search basis,
projected Hessian, and local metric are computed at the first
request and refreshed every $K$ requests thereafter.
At each refresh, the search subspace is constructed from the
current raw gradient and at most $r_{\max}-1$ recent gradients
stored in a history buffer of capacity $h$.
Numerically dependent directions are removed during orthogonalization.
HVPs are computed only if the
resulting rank satisfies $r_t\geq r_{\min}$. If the rank is too
small, another refresh is attempted at the next actor update. Other
invalid geometry results leave the stored geometry unusable until
the next scheduled refresh.

Between refreshes, the search basis $Q$, curvature eigenvectors
$U$, eigenvalues $\mu$, and metric weights $b$ remain fixed.
However, the raw gradient, matching scale, mixed gradient, and
selection checks are recomputed at every actor update.
Rejecting a mixed gradient does not trigger an extra refresh.
After a successful actor update, the history stores a detached
copy of the gradient $u_t$ supplied to the actor optimizer.

\subsection{\sacmethod\ implementation settings}
\label{app:settings}

We implement \sacmethod\ in Python by extending the PyTorch
implementation of Soft Actor--Critic (SAC) from OpenAI Spinning Up
(\url{https://github.com/openai/spinningup}).
Our implementation retains the original SAC training framework,
including the actor and critic networks, experience replay,
and Adam optimizers. The proposed \method\ components are
implemented separately and integrated into actor training
without modifying SAC's objectives for critic and actor update.
The SAC and PN baselines use the same underlying SAC codebase.

Our experiments use Python 3.10.20 and PyTorch 2.13.0
for neural network training, automatic differentiation,
and HVPs.
We use Gymnasium 1.3.0 to provide the MuJoCo-v5
continuous-control environments. The software environment
also includes NumPy 2.2.6 and SciPy 1.15.3.
Neural network computations use single-precision
floating-point arithmetic, while the eigendecomposition
of the projected Hessian is performed in double precision
to improve numerical stability.

Our experiments were conducted on CPU nodes of the
university's computing grid, with jobs managed through
the Sun Grid Engine. Python dependencies were managed using
Micromamba. The grid contains different CPU models, and
the computing resources were not identical across all runs.
Consequently, we evaluate learning performance primarily
in terms of environment interactions rather than
wall-clock training time.

\begin{table}[t]
\caption{Common SAC and \method\ settings.}
\label{tab:defaults}
\centering\small
\begin{tabular}{@{}p{0.46\linewidth}p{0.45\linewidth}@{}}
\toprule
Quantity & Setting\\
\midrule
\multicolumn{2}{l}{\emph{SAC}}\\
Actor and critic networks & Two 256-unit ReLU hidden layers\\
Actor and critic optimizers & Adam, learning rate $10^{-3}$\\
Discount factor and entropy temperature & $\gamma=0.99$, $\alpha=0.2$ (fixed)\\
Replay capacity and minibatch size & $10^6$ transitions, $100$\\
Initial random actions & $10^4$ environment interactions\\
Update schedule & Start after $10^3$ interactions; 50 updates per 50 interactions\\
Target-network Polyak retention & $0.995$\\
\midrule
\multicolumn{2}{l}{\emph{\method}}\\
Minimum subspace rank & $r_{\min}=4$\\
Curvature minibatches per refresh & $n_{\mathrm{curv}}=1$\\
Spectral weight bounds & $m=1/3$, $M=3$\\
Gradient mixing ratio & $\lambda=0.3$\\
Trace reduction threshold & $\rho_{\rm tr}=0.9$\\
Numerical regularizers & $\epsm=\epss=10^{-12}$\\
Request schedule & Every actor update\\
\bottomrule
\end{tabular}
\end{table}

Table~\ref{tab:defaults} summarizes the common SACand \method\ settings used in our experiments.
We evaluate the learned policies every 4,000 environment
interactions using 10 deterministic evaluation episodes.
The maximum episode length is 1,000 steps, subject to
the termination conditions of each benchmark task.
No additional observation or reward normalization is applied.
Separate random seeds are used for training environments,
evaluation environments, and replay sampling.

\section{Exploratory Extensions to EQUIP}
\label{app:exploratory-extensions}

We examine three extensions to \method\ that aim to improve actor
optimization through different gradient adjustment schedules or
curvature estimation strategies. These studies are conducted over 300k environment interactions and are separate from the main experiments
on the standard \sacmethod.

\subsection{Scheduled versus continuous gradient adjustments}
\label{app:exploratory-scheduling}

We first examine whether curvature-based gradient adjustments
should be requested throughout training or only at selected stages.
In our experiments on HalfCheetah, we compare two schedules. The
\emph{continuous} schedule requests an adjustment at every actor
update, subject to the usual eligibility and selection checks.
The \emph{scheduled} variant requests gradient adjustments during 20 actor
updates beginning at each of 120k, 160k, 200k, 240k, and 280k
environment interactions, for 100 requests in total.

Both variants follow identical hyper-parameter settings, including
$r_{\max}=8$, $h=16$, $K=50$, and $\lambda=0.3$.
Each is compared with the SAC baseline using the same five training
seeds ($0$--$4$). In our experiment, the minimum subspace rank is two
and the permitted candidate norm ratio is $[0.1,10]$. Hence, the
experiment results are not identical to the main benchmark runs.

\begin{table}[t]
\caption{Comparison of curvature-adjustment schedules on
HalfCheetah. Values show AUC and final-return differences
relative to SAC, with paired 95\% Student-$t$ confidence
intervals across five training seeds.}
\label{tab:app-scheduled-requests}
\centering\small
\setlength{\tabcolsep}{4pt}
\begin{tabular}{@{}lcc@{}}
\toprule
Request schedule & $\Delta$AUC [95\% CI]
                 & $\Delta$Return at 300k [95\% CI]\\
\midrule
Scheduled requests & $12.9\ [-9.8,35.7]$
                   & $-35.4\ [-326.2,255.4]$\\
Continuous requests & $257.5\ [7.6,507.3]$
                    & $1029.1\ [550.2,1507.9]$\\
\bottomrule
\end{tabular}
\end{table}

Table~\ref{tab:app-scheduled-requests} shows that continuous
requests achieve greater improvements over SAC in both
mean AUC and evaluation return at 300k.
In contrast, the scheduled variant provides little improvement.
Continuous requests also improve the mean return over the
last five evaluation checkpoints by $808.4$, with a paired
95\% confidence interval of $[428.4,1188.5]$.

\subsection{Hybrid spectral weighting}
\label{app:exploratory-hybrid}

We next test whether combining two spectral weighting rules
improves actor optimization. Specifically, the hybrid method uses the curvature-equilibration weights of \method\ when the projected Hessian
has robust positive and negative curvature. When the projected
Hessian is robustly positive definite, it instead uses the
inverse-absolute-curvature weights of PN
(Equation~\eqref{eq:exp-pn-weights}).

Both the hybrid and standard \sacmethod\ use the same hyper-parameter
settings, including $r_{\max}=32$, $h=32$, $K=64$, and $\lambda=0.3$.
As shown in Table~\ref{tab:app-hybrid}, the hybrid method achieves
a higher mean AUC on Walker2d ($1670.5$ versus $1537.5$).
Ant also shows a positive mean difference, whereas the other tasks do not.
Overall, the hybrid method does not consistently improve on the
standard \sacmethod. We therefore do not include it in the main
experiment results in Section~\ref{sec:exp-main}.

\begin{table}[t]
\caption{Hybrid spectral weighting compared with standard
\sacmethod\ under the same geometry configuration.
Values are paired AUC differences (hybrid minus standard)
over 300k environment interactions, with 95\%
Student-$t$ confidence intervals.}
\label{tab:app-hybrid}
\centering\small
\setlength{\tabcolsep}{4pt}
\begin{tabular}{@{}lrc c@{}}
\toprule
Task & Mean $\Delta$AUC [95\% CI] & Positive/negative\\
\midrule
HalfCheetah & $-84.7\ [-366.6,197.2]$ & $4/6$\\
Walker2d    & $133.0\ [-45.6,311.6]$ & $6/4$\\
Ant         & $35.4\ [-30.0,100.8]$ & $7/3$\\
Humanoid    & $-21.5\ [-42.9,0.0]$ & $5/5$\\
Swimmer     & $-0.041\ [-0.481,0.399]$ & $4/6$\\
Pusher      & $-0.632\ [-1.315,0.051]$ & $5/5$\\
\bottomrule
\end{tabular}
\end{table}

\subsection{Temporal averaging of projected Hessians}
\label{app:exploratory-temporal}

Finally, we investigate whether curvature information from
earlier stages of actor training can improve the estimation of
the current projected Hessian. Since the actor loss
changes during training, averaging projected Hessians
computed at different stages may reduce estimation
variance and provide more reliable curvature information
for actor optimization.

However, \method\ constructs its search subspace from
recent actor gradients, which evolve during training.
Consequently, projected Hessians computed at different
stages may correspond to different search subspaces and
cannot be directly averaged. We therefore develop a
basis-aligned temporal averaging mechanism.

Let $Q_t\in\mathbb R^{p\times r_t}$ denote the basis of
the current search subspace. At each geometry refresh, we
first compute the average projected Hessian over
$n_{\mathrm{curv}}$ curvature minibatches:
\begin{equation}
\widehat A_t =
\frac{1}{n_{\mathrm{curv}}}
\sum_{j=1}^{n_{\mathrm{curv}}}
Q_t^\top H_t^{(j)}Q_t,
\label{eq:temporal-current}
\end{equation}
where $H_t^{(j)}$ is the actor loss Hessian computed
from the $j$-th minibatch. All projected Hessians
use the same current search subspace.

Let $Q_{t^-}$ denote the basis of the search subspace at the
previous geometry refresh, and let $\widetilde A_{t^-}$
denote the projected Hessian estimate stored at that
refresh. Before combining this stored estimate with the
current estimate $\widehat A_t$, we express it in the current
search subspace coordinates:
\begin{equation}
\begin{aligned}
R_t = Q_t^\top Q_{t^-},
A_{t^-\rightarrow t}
=R_t\widetilde A_{t^-}R_t^\top.
\end{aligned}
\label{eq:temporal-transport}
\end{equation}
Here, $R_t$ relates the previous and current search bases,
while $A_{t^-\rightarrow t}$ represents the stored
curvature estimate in the current subspace coordinates.

Following the subspace-overlap measure used by
\citet{schneider2024subspaces}, we measure the overlap between
the two search subspaces as
\begin{equation}
o_t=\frac{\|Q_t^\top Q_{t^-}\|_F^2}{r_t},
\label{eq:temporal-overlap}
\end{equation}
where $\|\cdot\|_F$ denotes the Frobenius norm.
For subspaces of equal rank, $o_t$ ranges from zero
for orthogonal subspaces to one for identical subspaces.

When the two subspaces are sufficiently compatible ($o_t\geq 0.5$),
we combine the current and historical curvature estimates:
\begin{equation}
\widetilde A_t =
(1-\beta o_t)\widehat A_t
+\beta o_t A_{t^-\rightarrow t},
\label{eq:temporal-averaging}
\end{equation}
where $\beta\in[0,1]$ controls the strength of temporal
averaging. The historical estimate receives a weight
of $\beta o_t$, which decreases as the overlap between
the subspaces decreases.

Temporal averaging is applied only when the two subspaces
have the same rank and their overlap satisfies $o_t\geq0.5$.
Otherwise, we use the current projected Hessian estimate,
setting $\widetilde A_t=\widehat A_t$.
The resulting estimate is then used to construct the local
metric and adjust the actor gradient following the standard
\method\ procedure.

\paragraph{Experimental settings.}
We evaluate temporal averaging using a maximum subspace
rank of $r_{\max}=32$, history capacity $h=32$, refresh
interval $K=64$, and gradient mixing weight $\lambda=0.3$.
Each geometry refresh uses $n_{\mathrm{curv}}=4$
curvature minibatches. We examine three temporal
averaging weights, $\beta\in\{0.25,0.50,0.75\}$.
For comparison, we use the same four-minibatch
curvature estimator without temporal averaging.
All other settings remain unchanged.
The experiments cover 6 MuJoCo benchmarks using 10
training seeds per task, with performance evaluated over
300k environment interactions.

\paragraph{Experimental results.}
Compared with the four-minibatch estimator without
temporal averaging, the temporal variants improve mean
AUC on Pusher by $0.079$--$0.373$ across the three
values of $\beta$. However, no consistent improvements
are observed on the other benchmark tasks.
Overall, temporal averaging does not consistently improve
sample efficiency across the 6 benchmark tasks.

To understand these results, we examine how frequently
historical curvature estimates satisfy the conditions
required for temporal averaging. During 300k environment
interactions, none of the geometry refreshes on HalfCheetah,
Walker2d, or Ant satisfy these conditions. On Humanoid,
Swimmer, and Pusher, fewer than $0.2\%$ of refreshes allow
temporal averaging. Consequently, the algorithm almost
always relies on the current projected Hessian estimate.

These findings suggest that the trajectory-informed
search subspaces change substantially during actor training,
making historical curvature information difficult to reuse.
Given how rarely temporal averaging is applied, these
results do not demonstrate a clear benefit from reusing
historical curvature information.

\end{document}